\pdfoutput=1
\PassOptionsToPackage{table}{xcolor}

\documentclass[11pt]{article}

\usepackage[final]{acl}
\usepackage{listings}

\usepackage{times}
\usepackage{latexsym}

\usepackage[T1]{fontenc}

\usepackage[utf8]{inputenc}

\usepackage{microtype}

\usepackage{inconsolata}

\usepackage{graphicx}
 \usepackage{pgf}
\usepackage{transparent}
\usepackage{array}
\usepackage{multirow}
\usepackage{multicol}
\usepackage{xprintlen}
\usepackage{csquotes}
\usepackage[inline]{enumitem}
\usepackage[normalem]{ulem}
\usepackage[export]{adjustbox}
\usepackage{float}
\usepackage{placeins}
\usepackage{subcaption}
\usepackage{makecell}
\usepackage{tipa} %
\usepackage{svg}
\usepackage{xurl}
\usepackage[table]{xcolor}%
\usepackage{bbding}
\usepackage[outline]{contour}
\usepackage{tcolorbox}
\usepackage{amsmath,amsfonts,amssymb,amsthm}
\usepackage{lipsum}
\usepackage{siunitx}
\usepackage{tikz}
\usepackage{mathtools}
\usepackage{booktabs}
\usepackage{hyperref}
\usepackage{cleveref}
\usepackage[title]{appendix}
\usepackage{bbm}
\usepackage{soul}

  \everymath=\expandafter{\the\everymath\displaystyle}
  \makeatletter\@ifpackageloaded{underscore}{}{\usepackage[strings]{underscore}}\makeatother

\definecolor{teal}{RGB}{0,128,128}
\definecolor{darkpink}{RGB}{231,84,128}
\definecolor{darkgreen}{RGB}{50, 150, 50}

\newcolumntype{P}[1]{>{\centering\arraybackslash}m{#1}}
\newcolumntype{?}{!{\vrule width 1pt}}

\crefformat{section}{#2\S#1#3}
\Crefformat{section}{#2\S#1#3}
\crefformat{subsection}{#2\S#1#3}
\Crefformat{subsection}{#2\S#1#3}
\crefformat{subsubsection}{#2\S#1#3}
\Crefformat{subsubsection}{#2\S#1#3}
\crefformat{appendix}{#2App.~#1#3}
\Crefformat{appendix}{#2App.~#1#3}
\crefformat{figure}{#2Fig.~#1#3}
\crefformat{table}{#2Table~#1#3}

\newcommand{\lean}{\mathrm{lean}}
\newcommand{\dlean}{\Delta_{\lean}}
\newcommand{\drank}{\Delta_{\mathrm{rank}}}

\newcommand{\hcmsynth}{HCMagic$_{synth}$}
\newcommand{\hcmnat}{HCMagic$_{nat}$}
\newcommand{\allsidesq}{AllSides$_{synth}$}
\newcommand{\redditpol}{Reddit$_{nat}$}
\newcommand{\teLargeShort}{OpenAI \texttt{te-3-large}}

\definecolor{leftcolor}{HTML}{4477AA}
\definecolor{rightcolor}{HTML}{E60000}
\definecolor{wmecolor}{HTML}{EE7733}
\definecolor{aalcolor}{HTML}{AA3377}
\newcommand{\leftq}{{\color{leftcolor}left}}
\newcommand{\rightq}{{\color{rightcolor}right}}
\newcommand{\wmeq}{{\color{wmecolor}WME}}
\newcommand{\aalq}{{\color{aalcolor}AAL}}

\title{Retrieval Sensitivity to Identity Signals in Queries}

\author{
    \textbf{Andrew Tang}\textsuperscript{*},
    \textbf{Nicholas Deas}\textsuperscript{*},
    \textbf{Kathleen McKeown},
    \textbf{Vishal Misra}
    \\
    \\
    Columbia University, Department of Computer Science \\
    \small{
       Correspondence: \{a.tang, ndeas\}@cs.columbia.edu
     }
}

\begin{document}
\maketitle

\def\thefootnote{*}\footnotetext{Denotes equal contribution.}
\def\thefootnote{\arabic{footnote}}

\begin{abstract}
    Dense retrievers decide which documents reach users and the language models that use them, yet they are typically evaluated with neutral queries. We ask whether the identity signals that real users express in their queries---political ideology and dialect---bias what a retriever returns. We design evaluations in two domains, political news and consumer-health questions, each pairing a controlled synthetic set that varies only the identity signal with naturalistic queries. Across five dense retrievers and a sparse baseline, every retriever (i) retrieves articles that align with the query's own political lean and (ii) performs worse for questions written in African American Language (AAL) than in White Mainstream English (WME). Two analyses tie these gaps to queries' identity signals beyond surface vocabulary: partialling out an aggregate lexical-asymmetry score leaves the synthetic gaps largely intact, and linear probes recover lean and dialect from the retrievers' query embeddings beyond token-level features. Left unaddressed, such retrieval biases risk contributing to polarization and reinforcing the health disparities already faced by AAL speakers.\footnote{We make code available at  \url{https://github.com/Andrewtcr/bias-ret}.}
\end{abstract}

\newcommand{\x}{\mathbf{x}}
\newcommand{\y}{\mathbf{y}}
\newcommand{\xbar}{\overline{\x}}
\newcommand{\xhat}{\hat{\x}}
\newcommand{\bias}{\mathbf{b}}
\newcommand{\f}{\mathbf{f}}
\newcommand{\Fs}{\mathbf{F}}

\newcommand{\relu}{\text{ReLU}}
\newcommand{\ce}{\text{CE}}

\newcommand{\norm}[1]{\left\lVert#1\right\rVert}
\newcommand{\lex}{$\mathrm{lex}_q$}

\section{Introduction}
\label{sec:intro}

    Dense retrievers \cite{karpukhin-dpr} underpin search \cite{huang-ebr-search, zhao-dense-survey}, recommendation \cite{yi-twotower, pinterest-mer}, and retrieval-augmented generation (RAG) \cite{lewis-rag, fan-rag-survey}, selecting documents a user or downstream model sees. A retriever that systematically skews \emph{which} documents it surfaces is, therefore, consequential: in search and recommendation, the skewed results directly impact users, and in RAG, the skewed results are passed to a large language model (LLM), impacting its response.
    The latter is increasingly central as users shift toward AI-mediated information access \cite{genir-survey, liang-genir-users}, where poor or skewed retrieval sharply degrades generated outputs \cite{chen-rgb}.\footnote{Affected by generator robustness to imperfect context \cite{hsia-ragged}.}

    \begin{figure}[t!]
        \centering
        \includegraphics[width=0.95\linewidth]{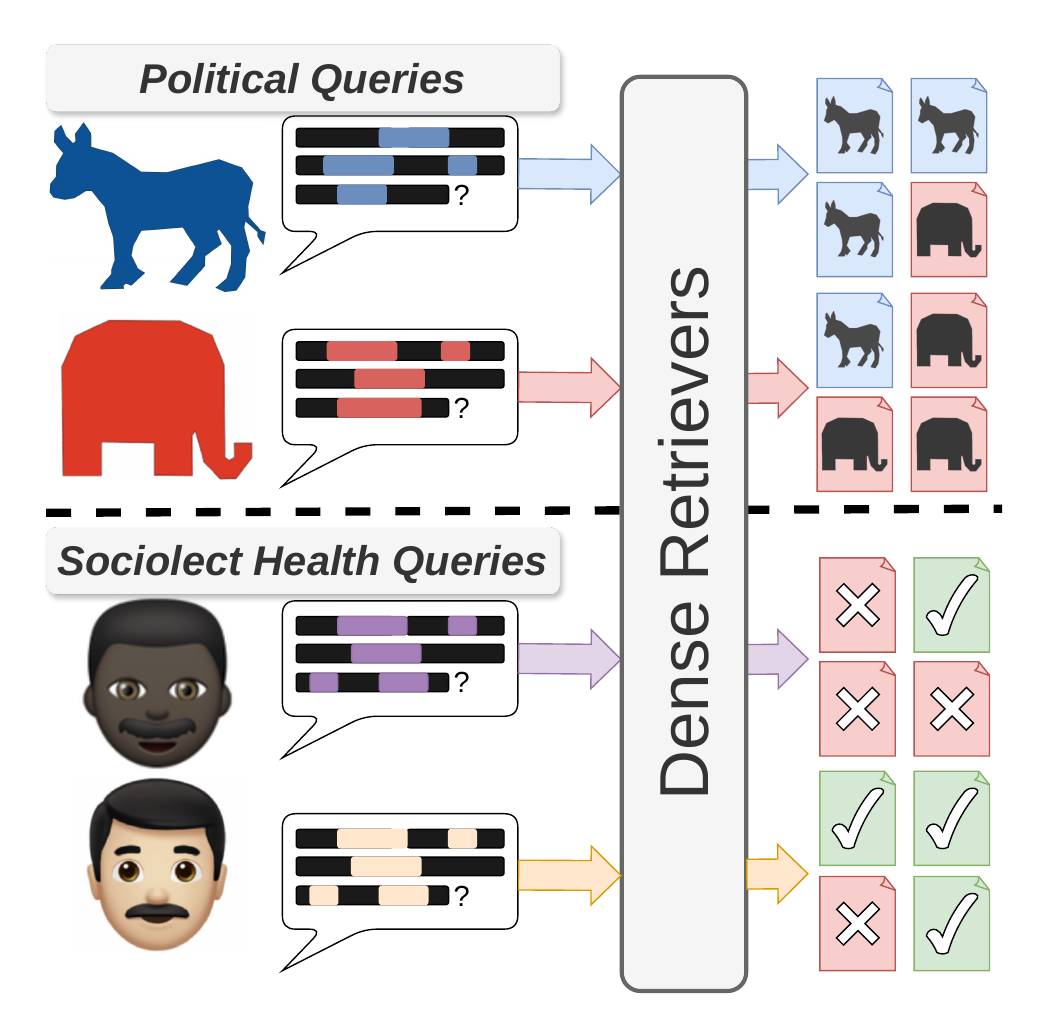}
        \caption{
        Paper overview.
        We evaluate retrievers' sensitivity to identity signals in queries, finding that retrievers favor documents aligning with the questioner's signaled political ideology (top) and that lower-quality documents are retrieved for AAL health queries (bottom).
        }
        \label{fig:teaser}
    \end{figure}

    Prior work shows that retrieval can be biased in two broad ways: by spurious features such as the position and length of documents \cite{yao-spotlight, zeng-position, cuconasu-rag}, and by
    social factors such as skewed gender representation
    among retrieved documents \cite{zhang-retrieval, parihar-social, krieg-bias}.
    \citet{kim-mitigating} frame the latter case as \emph{bias conflict}: corpora, query embeddings, and generators each carry their own bias, but the system's overall bias is not simply the sum of that of its components.
    Therefore, the
    retriever's own contribution is worth isolating.

    These studies, however, mostly examine relatively \emph{neutral} queries (queries that do not signal any particular sociodemographic group) and
    assess biases
    among retrieved documents (e.g., over-representation of one political ideology
    over another).
    Real queries instead routinely carry cues of the author's sociodemographics and ideology through different proxies (e.g., names, emotional reactions or values, other cultural proxies, etc.; \citealt{tonneau-cues, adilazuarda-towards}).
    Similarity-based retrieval makes some sensitivity to these cues expected in principle. This expectation, however, predicts neither which retrievers are affected, nor by how much, nor through which channel (surface vocabulary or the representation itself), and it assumes that documents aligned with the asker's identity are what they need.
    We therefore study an underexplored question: how \emph{identity signals in the query itself} lead a retriever to return systematically different documents.

    \begin{table*}[!t]
        \newlength{\polexw}
        \setlength{\polexw}{26cm}
        \centering
        \resizebox{.98\linewidth}{!}{
            \begin{tabular}{ccP{\polexw/3}P{\polexw/3}}
                 \toprule
                 & Frames & Left & Right \\
                 \midrule
                 \multirow{2}{*}{\rotatebox{90}{\allsidesq{}}} & \makecell{Fairness \\\& equality}
                  & \textit{Is raising the federal minimum wage to \$10.10 a {\color{leftcolor}\textbf{fair way to reduce inequality}} and {\color{leftcolor}\textbf{lift nearly a million from poverty}} per CBO?}
                  & \textit{Does the CBO report show raising the minimum wage to \$10.10 {\color{rightcolor}\textbf{unfairly distribute job losses}}, eliminating about 500,000 opportunities for workers?} \\
                  \cmidrule{2-4}
                  & Economic
                  & \textit{How would the CBO's \$10.10 minimum wage proposal economically {\color{leftcolor}\textbf{benefit low income workers}} and {\color{leftcolor}\textbf{reduce poverty}} despite some job losses?}
                  & \textit{What are the economic {\color{rightcolor}\textbf{costs to employment and small businesses}} of the CBO's \$10.10 minimum wage proposal, including projected job losses?} \\
                 \midrule
                 \rotatebox[origin=c]{90}{\redditpol{}}
                  & --- & \textit{Is anyone else {\color{leftcolor}\textbf{tired of
                    ground-level conservatives}} claiming they
                    {\color{leftcolor}\textbf{"agree"}} with news incidents while
                    always supporting the system that created these
                    injustices?}
                  & \textit{{\color{rightcolor}\textbf{Illegal aliens}} have always
                    been subject to deportation. Why does the government
                    now need the Alien Enemies Act to deport them?} \\
                 \bottomrule
            \end{tabular}
        }
        \caption{Select example queries from each political query dataset. Potential signals of the questioner's political ideology are highlighted {\color{rightcolor}\textbf{red}} and {\color{leftcolor}\textbf{blue}}.
        }
        \label{tab:data-exs}
    \end{table*}

    We demonstrate this in two domains closely related to downstream harms.
    First, we
    examine signals of the authors' political ideology (conveyed through presupposed beliefs \cite{sieker-presuppositions}, framing \cite{lee-neus}, or other means), where
 retrieval that aligns with the questioner's ideology may deepen polarization by reaffirming that ideology (\cref{sec:political}). Second, we test whether retrievers return lower-quality documents for health questions written in African American Language (AAL) than for those in White Mainstream English (WME), representing a direct \textit{allocational} harm\footnote{We consider this an \textit{allocational} harm because disparities in the performance of dense retrievers, particularly in the health domain, would provide inequitable access to informational health resources. See \citet{blodgett-language} for further discussion of allocational harms.} to AAL-speaking users (\cref{sec:aalwme}).\footnote{Prior work has used alternative terms, such as African American Vernacular English and Standard/Mainstream American English respectively. We follow recent work in using these terms to avoid implications of a ``standard variety'' \cite{baker-bell-aal,mckeown-aal}.}

    We detail our primary contributions as follows:
    \begin{enumerate*}[label=(\arabic*)]
        \item We build \textbf{query-based identity-bias evaluations for retrievers} for two variables---political ideology and AAL/WME sociolect---each relying on both a \emph{synthetic, paired} (i.e., topic-matched) query set that better isolates the identity signal, and a more \emph{ecologically valid, unpaired} query set of real user queries that better represents real-world deployments and potential harms.
        \item Evaluating \textbf{five dense retrievers} (and a sparse retriever baseline), we find that \textbf{every retriever prefers documents aligning with the questioner's political ideology} and \textbf{ranks the answer lower for AAL health queries than for WME}, consistent across the controlled and naturalistic settings.
        \item Two analyses tie these biases to the query's identity signal beyond
        surface vocabulary: on the content-controlled synthetic contrasts,
        \textbf{partialling out aggregate lexical asymmetry leaves the gap largely intact},
        and \textbf{linear probes recover lean/dialect from retriever query embeddings} beyond token-level features.
    \end{enumerate*}

\section{Political Bias}
\label{sec:political}

    We first ask whether dense retrievers produce systematically different
    news articles when provided queries signaling left and right political leanings.

    \subsection{Data}
    \label{sec:political-data}

        We evaluate dense retrievers on two complementary datasets of political queries: paired left and right-leaning queries about the same subject (\allsidesq{}) and unpaired naturalistic queries (\redditpol{}). Per-group query counts and length statistics are presented in \cref{app:data-stats}.

        \paragraph{Corpus.} We use the Qbias AllSides\footnote{\url{https://www.allsides.com/unbiased-balanced-news}} corpus \cite{haak-qbias} as the set of news articles to be retrieved.
        These articles are grouped into \textit{editorial roundups}, which are sets of articles about the same news story, but from sources with differing political leaning. Each grouping and article political leaning are determined by AllSides' cross-partisan team of expert news specialists.
        For evaluation, we extract a subset of Qbias for which the editorial roundups contain an equal number of left and right-leaning documents (${\sim}17$k articles total) to isolate retriever biases from corpus biases.

        \paragraph{\allsidesq{} (paired queries).}
        First, we consider a highly controlled setting of paired synthetic queries from $1{,}000$ randomly sampled roundups. For each roundup we use \texttt{gpt-5-mini} to label the Boydstun framing dimensions \cite{boydstun-mfc} present in its center article. For each frame, we then prompt the model to write two short search queries grounded in that article, one left-leaning and one right-leaning.
        Conditioning both queries on a shared frame is what makes the pair controlled: without it, a left- and a right-leaning asker may gravitate to different frames of the same story. In \cref{tab:data-exs}, a minimum-wage topic could draw a left asker toward the \textit{fairness \& equality} frame (reducing inequality, lifting people from poverty) and a right asker toward the \textit{economic} angle (costs to employment and small businesses)---the diagonal of the table.
        We generate every frame for both leans and measure the left/right contrast \emph{within} a frame.\footnote{\cref{app:pairing-check} confirms this design intent (L--R pairs from the same frame sit significantly closer in embedding space than L--R pairs from different frames, on every dense retriever).}
        This process results in $3{,}723$ left and $3{,}723$ right-leaning queries that share a common topic and frame with minimal differences
        (median length $138$ and $139$ characters, respectively).
        Prompts, the full Boydstun frame typology, and example frame-labeled articles are included in \cref{app:political-frames}.

        \paragraph{\redditpol{} (unpaired queries).}
        While the synthetic queries are paired and allow for controlled experiments, we additionally collect naturally occurring queries to better approximate realistic use cases.
        We collect $3{,}340$ posts from the \texttt{r/AskALiberal}\footnote{\url{https://www.reddit.com/r/AskALiberal/}} ($1{,}979$) and \texttt{r/AskConservatives}\footnote{\url{https://www.reddit.com/r/AskConservatives/}} ($1{,}361$) subreddits. These subreddits host community questions targeted toward liberal or conservative answerers, although the posters themselves may span the political spectrum. Following prior work \cite{tornberg-political}, we use LLMs to classify the ideology of a question's writer.
        Specifically, we pass the body of each post, which contains additional elaboration on each question, to
        two separate LLM classifiers,
 \texttt{gpt-5.4-mini}
        and \texttt{Qwen3.5-4B}, and keep only those where both agree.
        This yields $1{,}327$ liberal and $1{,}225$ conservative queries.
        Details on Reddit data collection and labeling are included in \cref{app:reddit-collection}.

    \subsection{Methods}
    \label{sec:political-methods}

        \begin{figure*}[!t]
            \centering
            \includegraphics[width=\linewidth]{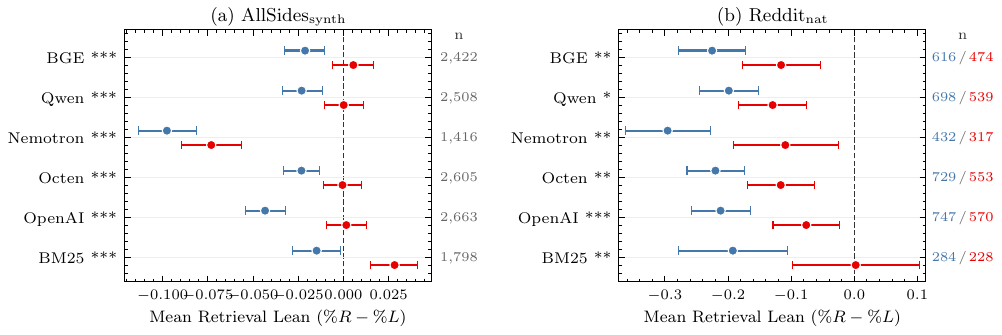}
            \caption{Per-retriever mean retrieval lean at $k{=}10$,
            gold-filtered (\cref{sec:political-methods}), shown separately for \leftq{} and \rightq{} query groups, with 95\%
            cluster-bootstrap CIs \citep{field-welsh-cluster-bootstrap}
            ($B{=}5{,}000$ resamples; clusters $=$ queries from the same
            source article).
            Own-lean bias appears as the displacement between a
            retriever's two markers; an unbiased retriever's would
            coincide. The markers' shared offset from zero reflects a
            corpus-level baseline common to both query groups.
            Stars mark the significance of the own-lean gap: \texttt{*}$p{<}0.05$,
            \texttt{**}$p{<}0.01$, \texttt{***}$p{<}0.001$.
            $n$ queries remaining after gold-filtering, in the right
            margin. Per-retriever gaps in \cref{app:political-goldtest}.}
            \label{fig:political-combined}
        \end{figure*}

        \paragraph{Retrievers.}
        We evaluate BGE-large-en-v1.5 \cite{xiao-bge}, Qwen3-Embedding-8B
        \cite{zhang-qwen3}, Llama-Nemotron-8B \cite{babakhin-embedding}, and
        Octen-Embedding-8B \cite{octen-rteb} encoders for dense retrieval.\footnote{Implementation details in \cref{app:model-details}.}
        Selection was based on high performing retrievers on the MTEB benchmark \cite{muennighoff-mteb} at the time of writing while representing different families of open-source models. We also include OpenAI's
        \texttt{text-embedding-3-large} (\teLargeShort{})
        and
        BM25-Okapi \cite{robertson-bm25}
        as closed-source and sparse retrieval (lexical) baselines, respectively.

        \paragraph{Political lean metric.} To evaluate biases, we focus on the representation of left and right-leaning articles among retrieved documents. Specifically, we measure the political lean of a query's top-$k$ retrieval set as the difference between the fraction of right- and left-leaning articles it contains:
        $$\lean(q) = \frac{1}{k}\left(|d^{(k)}_R| - |d^{(k)}_L|\right) \in [-1, +1],$$
        where $d^{(k)}$ is the set of top-$k$ retrieved articles and the
        subscript denotes the corresponding lean subset. Values near $+1$
        indicate a predominantly right-leaning set and values near $-1$ a
        left-leaning one, with center articles excluded from the count.

        \paragraph{Gold-filtering.}
        To measure lean over the documents that actually engage with a query rather than off-topic retrievals, we define a \emph{gold} (relevant) set per query.
        On \allsidesq{}, a query's gold articles are the others in its source editorial roundup (the source center article, which contains the answer, is masked during retrieval). We keep a left/right pair when \emph{both} queries retrieve a gold article and measure lean over the full top-$k$. On \redditpol{}, which has no gold, we LLM-judge every (query, article) pair in the top-$k$ with two independent judges (\texttt{gpt-5-mini} and \texttt{Qwen3.5-35B-A3B}) and count an article gold only when \emph{both} grade it relevant (binary; \cref{app:reddit-judge}), measuring lean over the relevant articles (normalized by the relevant count rather than $k$). This filtering is conservative. With no relevance filter at all, the lean
        gap is no smaller---and larger for most retrievers---and it persists across other filter settings we examine in \cref{app:gold-filter}.

        \paragraph{Significance tests.}
        We expect an unbiased retriever to similarly represent left and right documents in the retrieved set, regardless of the query's leaning. Therefore, we need to know whether any \emph{own-lean tendency} (the tendency for a query's top-$k$ to be shifted toward articles labeled with the same lean as the query) is statistically distinguishable from no lean preference (i.e., no bias).

        For \allsidesq{}, we formally test the paired lean difference
        between right and left queries from the same source article and frame,
        $$\dlean = \lean(q^{(R)}) - \lean(q^{(L)}),$$
        computed over gold-filtered query pairs. We summarise this
        contrast by $\overline{\dlean}$, the mean
        of $\dlean$ across
        pairs; $\overline{\dlean} > 0$ indicates that a retriever is
        biased toward documents with the same political lean as the original query.\footnote{$\Delta_{\lean} \in [-2, +2]$, where $+2$ means right queries retrieve right articles only and left queries retrieve left articles only, and vice versa.}
        We test $\overline{\dlean}$ against the null hypothesis of no lean preference (i.e., $\overline{\dlean} = 0$). For \redditpol{}, queries are unpaired, so we instead test
        whether left query $\lean$ is systematically lower than right
        query $\lean$.\footnote{
            \textit{Sign-flip
            permutation} \citep{pitman-permutation} and two-sided
            \textit{paired Wilcoxon signed-rank} test
            \citep{wilcoxon-signedrank} for paired, two-sided \textit{Mann--Whitney $U$} \citep{mann-whitney} for unpaired. Details in \cref{app:sigtests}.
        }

    \subsection{Results}
    \label{sec:political-results}

        \paragraph{Own-lean tendency on both datasets.}
        On both \allsidesq{} and \redditpol{}, every retriever shows significant
        own-lean tendency (\cref{fig:political-combined}; both subfigures
        reject the null hypothesis of no lean preference on every retriever at
        $p<0.05$; per-retriever own-lean contrasts with CIs and $p$ for
        both datasets in \cref{app:political-goldtest}).
        For \allsidesq{}, BM25 and OpenAI have about $2\times$ the gap of the four open-weight models. For \redditpol{}, BM25 and Nemotron have the largest gaps (about $2\times$ that of BGE, Qwen, and Octen), with OpenAI in-between.
        This own-lean tendency is present at retrieval depth $k\in\{5,10,20,50,100\}$
        (see \cref{app:political-ksweeps}).

        \begin{table*}[!t]
            \newlength{\dialexw}
            \setlength{\dialexw}{19cm}
            \centering
            \resizebox{.95\linewidth}{!}{
                \begin{tabular}{cP{9cm}P{10cm}}
                     \toprule
                     & AAL & WME \\
                     \midrule
                     \hcmsynth{}
                       & \textit{Hi {\color{aalcolor}docta}, how
                         {\color{aalcolor}might can} that renal
                         {\color{aalcolor}functions} be monitored in
                         {\color{aalcolor}one} patient undergoing
                         hemodialysis 3 times per week?}
                       & \textit{Hi doctor, how can the renal function
                         be monitored in a patient undergoing
                         hemodialysis 3 times per week?} \\
                     \midrule
                     \hcmnat{}
                       & \textit{...about 2 weeks ago she {\color{aalcolor}notice}
                         she had a lump just below her left rib in the
                         front\ldots While standing if she twist her body
                         you can feel it but once {\color{aalcolor}ahe straight}
                         again u can see it and feel it. What can this be.}
                       & \textit{Every now and then I will feel my chest
                         tighten and my heart starts pounding extremely
                         hard. I cant breath and I begin feeling very
                         lightheaded. It usually only lasts a few seconds
                         and does not happen too often. Any ideas as to
                         what could be causing this?} \\
                     \bottomrule
                \end{tabular}
            }
            \caption{Select example queries from each AAL/WME
            dataset. AAL surface markers are coloured
            {\color{aalcolor}purple}.}
            \label{tab:aalwme-exs}
        \end{table*}

        \paragraph{Paired vs unpaired contrast.}
        The agreement between the paired \allsidesq{} and unpaired \redditpol{} tests strongly suggests political-identity signals in the query are the source of retrieval bias. Because the paired queries hold topic and frame fixed (\cref{sec:political-data}), \cref{fig:political-combined}a isolates the gap to the queries' political identity rather than to the story or frame a partisan asker would choose. The naturalistic agreement in \cref{fig:political-combined}b then rules out synthetic-pipeline artifacts.

        \paragraph{Dense gaps do not require token overlap.}
        How much vocabulary a query shares with the corpus varies sharply across our two settings, and BM25's own-lean tendency is affected by it while the dense retrievers' is not. BM25 has among the highest own-lean gaps on both \allsidesq{} and \redditpol{}, but it has no gap for \redditpol{} without the gold-filter, as its unfiltered top-10 is dominated by off-topic documents (\cref{app:gold-filter}). In contrast, dense retrievers' gaps remain significant (\cref{app:gold-filter}), indicating that dense retrievers' own-lean tendency does not require exact token matching. Whether surface vocabulary or the representation itself carries the signal is investigated in \cref{sec:bias-gap}.

\section{AAL vs.\ WME in Health Queries}
\label{sec:aalwme}

    We then test whether dense retrievers' performance degrades on African American Language
    (AAL) consumer-health queries relative to White Mainstream English
    (WME).

    \subsection{Data}
    \label{sec:aalwme-data}

    We pair a synthetic dataset (\hcmsynth{})
    with an unpaired naturalistic one (\hcmnat{}) over a shared healthcare
    question-answering corpus (counts in \cref{app:data-stats}), as in \cref{sec:political}. We study
    health because bias or performance disparities in retrieval here can carry severe real-world
    consequences, as documented for other language technologies
    (e.g., \citealt{harrington-health}).

        \paragraph{Corpus.}
        We draw both queries and retrieval documents from the ChatDoctor HealthCareMagic-100k corpus \cite{li-chatdoctor}, an online platform where users of any background ask questions about their health and receive responses from affiliated doctors. \citet{li-chatdoctor} curate $112{,}165$ user question and doctor response pairs. We index every doctor response as a corpus of retrieval documents. For a given patient query (\cref{tab:aalwme-exs}) the task is to surface the doctor response (gold document) that answers it.
        \cref{app:hcmagic-corpus-ex} gives query--document examples, with the gold doctor response, for both datasets. The two query datasets below both sample from this corpus: \hcmsynth{} from questions classified as WME (with AAL added synthetically) and \hcmnat{} from naturally-occurring AAL and WME questions.

        \paragraph{\hcmsynth{} (paired queries).}
        We sample $4{,}999$ HealthCareMagic posts\footnote{We originally sample 5,000, but one sample is dropped due to an empty string resulting from AAL augmentation.} whose patient question is classified as WME by the demographic-alignment classifier of \citet{blodgett-demographic}.
        We synthetically augment each WME query with AAL features using the methods of \citet{ziems-value} (morphosyntactic) and \citet{deas-phonate} (phonological/orthographic). Although imperfect, both of these approaches were validated with human judgments in their corresponding studies and have been used to similarly evaluate other language technologies (e.g., reward models; \citealp{mire-dialects}).

        As these methods are non-deterministic, we generate three independent augmentations of each query to account for variability, yielding $4{,}999\!\times\!3{=}14{,}997$
        AAL queries. In using this data for our evaluations
        (\cref{sec:aalwme-methods}), we therefore hold content fixed while varying dialectal features.

        \paragraph{\hcmnat{} (unpaired queries).} For realistic queries, we use the same demographic-alignment classifier \cite{blodgett-demographic} to surface queries that use features associated with AAL, resulting in $656$ AAL queries ($647$ after filtering). We randomly sample a similar number of real-WME queries.\footnote{Queries are filtered to $\ge20$ characters, details in \cref{app:hcmagic-filter}.}

        \begin{figure*}[!t]
            \centering
            \includegraphics[width=\linewidth]{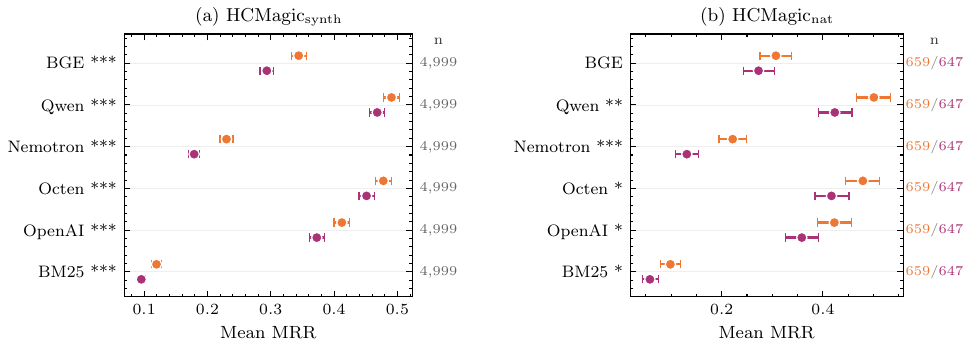}
            \vspace{-2em}
            \caption{Per-retriever mean MRR on the shared
            HealthCareMagic corpus, shown separately for \wmeq{} and \aalq{} queries, with 95\%
            bootstrap CIs.
            Stars mark the significance of the WME-AAL gap: \texttt{*}$p{<}0.05$,
            \texttt{**}$p{<}0.01$, \texttt{***}$p{<}0.001$. $n$ queries in the
            right margin. Per-retriever gaps in \cref{app:hcmagic-real}.}
            \label{fig:aalwme-combined}
        \end{figure*}

\subsection{Methods}
    \label{sec:aalwme-methods}

        \paragraph{Retrievers.} For AAL/WME experiments, we use the same six retrievers as
        \cref{sec:political-methods}.

        \paragraph{Retrieval metric.}
        We score each query by reciprocal rank, $1/r$, where $r$ is
        the rank of the gold \texttt{output} document in the
        retriever's full sorted ranking over the $112{,}165$-document
        corpus, following the standard retrieval-QA protocol of
        treating a question's original paired answer as the gold document
        \citep{ahmad-reqa, guo-multireqa}. Mean reciprocal rank (MRR) is the
        mean of $1/r$ across a query set. Unlike \cref{sec:political-methods},
        we do not gold-filter to relevant documents because retrieval competence (MRR) is the metric of interest.

        \paragraph{Significance tests.}
        For \hcmsynth{}, the sample-level paired difference is
        $\drank = 1/r^{\text{WME}}_r - \overline{1/r^{\text{AAL}}_r}$
        (WME reciprocal rank minus the mean over the sample's three synthetic AAL
 rewritings), summarised across samples by $\overline{\drank}$. This parallels the per-pair lean difference $\dlean$ of
        \cref{sec:political-methods}.
        For \hcmnat{}, queries are unpaired, so we report the between-group
        MRR difference $\mathrm{MRR}_{\text{WME}} - \mathrm{MRR}_{\text{AAL}}$,
        tested with a two-sided Mann--Whitney $U$ (paired Wilcoxon and
        sign-flip permutation for \hcmsynth{}; see \cref{app:sigtests}).

    \subsection{Results}
    \label{sec:aalwme-results}

        \paragraph{WME${>}$AAL on both datasets.}
        On both \hcmsynth{} and \hcmnat{}, every retriever shows
        $\mathrm{MRR}_{\text{WME}} > \mathrm{MRR}_{\text{AAL}}$
        (\cref{fig:aalwme-combined}; full per-retriever numbers
        in \cref{app:hcmagic-real}).
        On \hcmsynth{} the paired Wilcoxon rejects on all
        six retrievers ($p\!\approx\!0$). On \hcmnat{}, five of six
        reject at $p<0.05$ (BGE marginal at
        $p{=}0.10$).
        Using AAL rather than WME for the same health question consistently lowers the rank of its
        answer document, so retrievers serve questioners using AAL worse than those using WME
 with near identical information need. The gap is stable across
        retrieval depth $k$ (\cref{app:aalwme-ksweeps}).

        \paragraph{Paired vs unpaired contrast.}
        The two datasets agree on the harm. Because \hcmsynth{} holds
        content fixed and varies only dialect (each WME query paired against its own AAL rewritings), its gap shows that features associated with AAL
        \emph{alone} lower retrieval quality in this domain and attributes the direction
        of the effect (WME above AAL) to dialect rather than any content
        confound. We note, however, that on its own, it is likely an overestimate of the dialect effect,
        since the synthetic rewrites add some noise
        (\cref{app:paraphrase-qc}). \hcmnat{} instead compares different
        patient pools, mixing dialect with real differences in content and
        topic difficulty, and its gap tends to be $\sim1.7\times$ larger for all retrievers.
        This indicates that dialect by itself causes the
        disparity, and in real deployments---where AAL and WME queries
        differ in content as well as dialect---the disadvantage AAL users
        face may be larger still.\footnote{We also assess a simple deployment mitigation, back-translating each AAL query to WME, but it does not close the gap with WME and only helps weaker retrievers (\cref{app:translation}).}

\section{Bias gap analysis}
\label{sec:bias-gap}
\label{sec:political-lexreg-main}
\label{sec:aalwme-lexreg-main}

    What drives the retrieval gaps in \cref{sec:political-results} and
    \cref{sec:aalwme-results}? Left and right queries, like AAL and WME
    queries, differ in surface vocabulary (lean-based word choices, dialect morphology and respellings), so a retriever that mainly matches
    surface tokens could reproduce the gap without encoding the query's
    political lean or dialect at all (the \emph{lexical} component).
    Alternatively, the embedding may capture lean or dialect identity beyond surface tokens (through morphosyntax, composition, or semantics), shifting
    retrieval even when vocabulary is held fixed (the \emph{non-lexical}
    component). We separate these with two analyses: partialling out a per-query lexical-asymmetry score, and probing each dense retriever's query embeddings for whether lean or dialect is linearly encoded beyond token-level features. The first is a conservative test: the groups' vocabulary differences may themselves reflect bias, so a surviving gap is specifically attributable to the retriever rather than to vocabulary.

    \subsection{Method}
    \begin{table*}[!t]
            \centering\small
            \setlength{\tabcolsep}{5pt}
            \begin{tabular}{l S S@{}l S S@{}l S S@{}l S S@{}l}
                \toprule
                & \multicolumn{3}{c}{\allsidesq{} ($\dlean$)}
                & \multicolumn{3}{c}{\redditpol{} (lean)}
                & \multicolumn{3}{c}{\hcmsynth{} ($\drank$)}
                & \multicolumn{3}{c}{\hcmnat{} (MRR)} \\
                \cmidrule(lr){2-4}\cmidrule(lr){5-7}\cmidrule(lr){8-10}\cmidrule(lr){11-13}
                Retriever & {raw} & \multicolumn{2}{c}{resid.} & {raw} & \multicolumn{2}{c}{resid.} & {raw} & \multicolumn{2}{c}{resid.} & {raw} & \multicolumn{2}{c}{resid.} \\
                \midrule
                BGE-large         & +0.027 & +0.027 & $^{***}$ & +0.024 & +0.007 & {}        & +0.050 & +0.047 & $^{***}$ & +0.034 & +0.042 & {} \\
                Qwen3-Emb-8B      & +0.023 & +0.031 & $^{***}$ & +0.028 & +0.015 & {}        & +0.023 & +0.021 & $^{***}$ & +0.077 & +0.110 & $^{**}$ \\
                Llama-Nemotron-8B & +0.025 & +0.025 & $^{*}$   & +0.039 & +0.022 & $^{*}$    & +0.051 & +0.046 & $^{***}$ & +0.091 & +0.085 & $^{**}$ \\
                Octen-Emb-8B      & +0.023 & +0.018 & $^{**}$  & +0.048 & +0.033 & $^{***}$  & +0.027 & +0.024 & $^{***}$ & +0.062 & +0.098 & $^{**}$ \\
                \teLargeShort{}   & +0.045 & +0.029 & $^{***}$ & +0.062 & +0.036 & $^{***}$  & +0.040 & +0.039 & $^{***}$ & +0.064 & +0.081 & $^{*}$ \\
                \midrule
                BM25 (Okapi)      & +0.043 & +0.038 & $^{***}$ & +0.001 & +0.002 & {}        & +0.024 & +0.024 & $^{***}$ & +0.040 & +0.041 & $^{*}$ \\
                \bottomrule
            \end{tabular}
            \caption{Per-retriever \emph{raw} bias gap and \emph{residual}
            gap after partialling out the lexical channel (definitions in
            \cref{sec:bias-gap-method}).
            Stars: ${}^{*}p{<}0.05$, ${}^{**}p{<}0.01$, ${}^{***}p{<}0.001$.
            Stars test the \emph{residual} against zero
            ($\hat\alpha$ for the paired datasets, $\hat\gamma$ for the
            unpaired); an unstarred residual is indistinguishable from
            zero. Raw columns carry no stars because they are not the
            quantity under test here: they correspond to the gaps of
            \cref{sec:political-results} and \cref{sec:aalwme-results}.
            Full coefficients
            in \cref{app:lexreg}.
            }
            \label{tab:bias-gap}
        \end{table*}
    \label{sec:bias-gap-method}

        \paragraph{Scoring a query's lexical asymmetry.}
        We need a per-query measure of how strongly the query's
        vocabulary leans toward one group or the other. We use the
        MCQ log-odds asymmetry
        \citep{monroe-fightin-words}, a standard
        computational social science tool for surfacing the words that
        distinguish two sides of a corpus. MCQ assigns each token a
        score, $\zeta$, where large positive $\zeta$ indicates tokens used
        disproportionately by one group, large negative $\zeta$ indicates those disproportionately used by the
        other group, and near zero indicates tokens used equally.\footnote{Note that zero here does not distinguish between equal number of extremely right/left tokens and all "neutral" tokens.}
        Summing $\zeta$
        over the tokens in a query gives its \emph{lexical asymmetry},
        \lex{}. For political
        datasets we fit $\zeta$ on the corresponding \leftq{}-vs-\rightq{} query corpora;
        for dialect datasets, on the
        \aalq{}-vs-\wmeq{} query corpora.

        \paragraph{Controlling for lexical asymmetry.}
        We ask: \textit{how much of the bias gap is explained by \lex{}}, and
        \textit{how much remains after we subtract the \lex{} contribution}? We
        investigate these questions by regressing the gap on \lex{}. The regression intercept is the bias that \lex{} cannot
        explain, and the regression slope tells us how strongly the
        gap moves with lexical asymmetry. We call this
        \emph{partialling out lexical asymmetry}.\footnote{Tokenization and
        standard-error details in \cref{app:bias-gap-details}.}
        On the two paired, synthetic datasets (\allsidesq{} and \hcmsynth{}),
        we regress on the paired \lex{} difference:
        $$
        \Delta_q = \alpha + \beta\,\bigl(\mathrm{lex}_q^{(1)} -
        \mathrm{lex}_q^{(2)}\bigr) + \varepsilon_q
        \quad\text{[paired]},
        $$
        where $\Delta_q$ is $\dlean = \lean(R) - \lean(L)$ on
        \allsidesq{} (groups $1{=}$R, $2{=}$L) and $\drank$, the per-sample
        reciprocal-rank difference, on \hcmsynth{} (groups $1{=}$WME,
        $2{=}$AAL; $\mathrm{lex_q}^{(2)}$ averaged over the sample's three
        AAL rewritings). We reserve $\Delta$ for these paired differences. The unpaired gaps below are not written as $\Delta$. On the two unpaired datasets, the gap is a
        between-group measure, so we regress the per-query outcome
        on per-query \lex{} plus a binary group indicator:
        $$
        Y_q = \alpha + \beta\,\mathrm{lex}_q +
        \gamma\,\mathbbm{1}\!\left[q \in G_2\right] + \varepsilon_q
        \quad\text{[unpaired]},
        $$
        with $Y_q$ the per-query $\lean(q)$ on \redditpol{} ($G_2 =$
        right queries) and the per-query reciprocal rank $1/r$ on
        \hcmnat{} ($G_2 =$ WME pool).

        \paragraph{What the coefficients mean.}
        The \textbf{raw gap} is the unadjusted bias gap measured in
        \cref{sec:political-results} and \cref{sec:aalwme-results},
        with no \lex{} term in the model: the mean paired difference
        ($\overline{\dlean}$ or $\overline{\drank}$) on the paired datasets and the raw
        between-group difference in $Y$ (e.g.\ mean
        $\lean$ of right minus left queries) on the unpaired datasets.
        The \textbf{residual gap} is the part of that gap
        that is conserved after controlling for \lex{}: $\hat\alpha$ in the paired
        case (the expected paired difference at zero per-pair \lex{} difference)
        and $\hat\gamma$ in the unpaired case (the expected
        between-group difference in $Y$ at zero per-query \lex{}). If
        \lex{} perfectly explained the gap, the residual would be zero;
        if \lex{} explained nothing, it would equal the raw gap. The
        \textbf{\lex{} slope}, $\hat\beta$, quantifies how much of the
        gap moves with \lex{}: a non-zero $\hat\beta$ means the
        retriever's score tracks per-query \lex{} within or across the
        groups, whereas $\hat\beta \approx 0$ means \lex{} is not a useful
        predictor of the gap.

        \paragraph{Conditional probing.}
        On the two paired synthetic datasets, we fit linear probes
        (logistic regression) on each open-weight dense retriever's mean-pooled query
        embedding at every layer, predicting \leftq{}/\rightq{}
        (\allsidesq{}) or \wmeq{}/\aalq{} (\hcmsynth{}). We report probe F1
        (class separability) and, following \emph{conditional probing}
        \citep{hewitt-conditional}, $\mathcal{V}$-information, the gain in
        negative log-loss over a baseline probe trained on the model's
        non-contextual embedding layer. In other words, this measure is a proxy for the social identity information a model layer adds \emph{beyond} token-level features alone (positive meaning more information). Evaluation uses
        5-fold cross-validation (political) or a held-out $20\%$ split
        averaged over the three AAL augmentations (dialect). Additional details included in \cref{app:probing-dets}.

        \begin{figure}[H]
            \centering
            \includegraphics[width=0.95\linewidth]{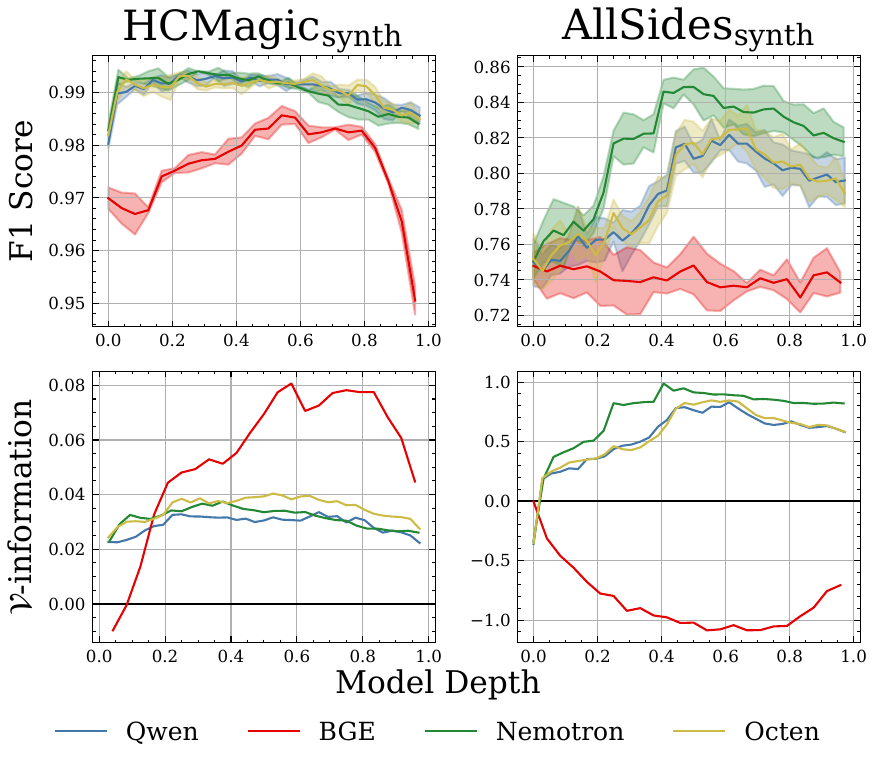}
            \vspace{-1em}
            \caption{Average F1 score (top) and $\mathcal{V}$-information (bottom) for normal and conditional probes, respectively, at each model depth. F1 indicates how readily lean or dialect can be decoded at each layer; $\mathcal{V}$-information indicates what that layer adds beyond the input embeddings. Both signals remain decodable across depth, with positive conditional gains at most layers except for BGE on AllSides.}
            \vspace{-1em}
            \label{fig:synth-probe-nce}
        \end{figure}

    \subsection{Results}
    \label{sec:bias-gap-results}

        \paragraph{Synthetic datasets: the gap survives lexical-asymmetry control.}
        On both \allsidesq{} and \hcmsynth{}, partialling out \lex{}
        does not substantially change the gaps across retrievers: the residual stays close to
        the raw gap and remains significant on every retriever
        (\cref{tab:bias-gap}). The only retriever with a detectable
        \lex{} slope is \teLargeShort{} on \allsidesq{}, where partialling
        absorbs the part of its gap that exceeds the open-weights but
        leaves a retriever-level residual comparable to theirs. BM25
        carries one of the largest \allsidesq{} residuals, so under
        content control the lexical channel itself is at least as
        biased as the dense retrievers.

        \paragraph{Naturalistic datasets: \lex{} partials some
        retrievers, not others.}
        On \redditpol{}, partialling out \lex{} shrinks every retriever's
        gap, but by enough to change the conclusion for only some
        (\cref{tab:bias-gap}). For the two dense retrievers with the
        smallest raw gaps (BGE, Qwen3) the residual shrinks into
        statistical insignificance, so their naturalistic gap is
        consistent with topic-asymmetry exposure rather than a
        retriever-level non-lexical component. BM25 has no \redditpol{} gap to begin with. The larger-gap dense retrievers (Octen,
        OpenAI, Nemotron) retain a significant residual,
        so their naturalistic bias has a component per-query \lex{} does
        not absorb.
        On \hcmnat{}, partialling does not shrink the gap into insignificance on any retriever. The residual stays close to
        or \emph{above} the raw gap. This is a confounding
        artifact, not a deeper-than-lexical signal. The
        ``AAL-distinctive'' tokens that MCQ picks up here are
        dominated by patient-topic words (\textit{she}, \textit{her},
        \textit{periods}, \textit{baby}), reflecting demographic
        differences between the natural AAL and WME query pools
        rather than dialect surface form, so the \hcmnat{} residual is
        not a clean non-lexical measure. The translation
        intervention (\cref{fig:aalwme-translation-forest})
        remains the proper content-difficulty probe on \hcmnat{}.

        \paragraph{What this tells us.}
        The synthetic-vs-naturalistic split confirms the design
        intent of the synthetic contrasts. By holding content fixed
        (\allsidesq{}'s left and right queries are written about the
        same source article and frame; \hcmsynth{}'s AAL queries are
        rewritings of the same WME source), the synthetic contrasts
        target the gap contribution that is \emph{not} driven by
        lexical token frequency, isolating non-lexical differences
        in how the retriever represents lean or dialect identity. The
        naturalistic contrasts mix these non-lexical differences with
        lexical and topical differences across the two query pools,
        so partialling out \lex{} absorbs the gap on retrievers
        whose naturalistic bias is largely token-driven (BGE and Qwen3
        on \redditpol{}).
        Full per-retriever tables in \cref{app:lexreg}.

        \paragraph{The signal is present beyond the lexical channel.}
        Probing provides complementary results: lean and dialect are linearly
        recoverable from the dense retrievers' query embeddings (probe F1
        ${\approx}\,85\%$ and $99\%$; \cref{fig:synth-probe-nce}), and
        conditional probing shows that, for most layers, additional identity information is encoded \textit{beyond} the non-contextual baseline.
        This is a prerequisite for, not
        proof of, a non-lexical gap: the signal a retriever could match on
        is present and not purely lexical.

\section{Related Work}

    \paragraph{Identity signals in language models.}
        Encoders represent authors' sociodemographic attributes such as
        gender and age \cite{lauscher-socioprobe}, and LLMs both misunderstand
        AAL features \cite{deas-aal, fleisig-dialect, mckeown-aal} and
        stereotype its speakers \cite{deas-data, deas-phonate, hofmann-dialect}.
        They are likewise sensitive to political cues: presupposed beliefs
        \cite{sieker-presuppositions}, framing \cite{lee-neus}, and named
        entities in sentiment tasks
        \cite{plisiecki-political, ng-stance}. We study these signals carried
        \emph{in the query} and how they steer a retriever.

    \paragraph{Bias in retrieval and RAG.}
        Retrieval and RAG are skewed by spurious features such as evidence
        position and length \cite{zeng-position, liu-lost, cuconasu-rag} and by
        social bias in the returned set: rankers reinforce gender bias
        \cite{rekabsaz-societal}, benchmarks quantify representation bias over
        neutral queries \cite{krieg-bias}, and fair-ranking work targets
        equitable \emph{exposure} across document groups
        \cite{singh-fairexposure, zehlike-fair}. As retrieval quality bounds
        generation \cite{chen-rgb, hsia-ragged}, this bias propagates through
        RAG \cite{kim-mitigating, wu-rag-fairness}. A separate line treats query \emph{phrasing} (misspellings and reformulations) as noise to be
        robust to \cite{penha-robustness, sidiropoulos-misspellings}. Closest to
        us, \citet{kim-mitigating} (political bias) and \citet{klisura-dialect}
        (dialect QA) still use neutral or scenario queries, or measure
        end-to-end accuracy. We instead make the query's identity signal the
        variable of interest, measuring how the retrieved set shifts with the
        asker's lean or dialect.

\section{Conclusion}
\label{sec:conclusion}

    Dense retrievers are sensitive to identity signals carried in the query
    itself. Across five dense retrievers and a sparse baseline, retrieval
    shifts toward the questioner's own political lean and ranks answers lower
    for AAL than WME health queries, in both a content-controlled paired
    design and naturally occurring queries.
    On the content-controlled synthetic contrasts the gap survives partialling
    out aggregate lexical asymmetry, and lean and dialect are linearly
    recoverable from the query embeddings beyond token-level features, so the
    bias tracks the query's identity signal, not merely its surface vocabulary.
    These shifts risk contributing to political polarization as well as health
    disparities faced by AAL speakers. Consistent across architectures and
    not closed by back-translation, they likely require representation-level
    mitigation, which we leave to future work.

\section*{Limitations}

    \paragraph{Coarse operationalization of identity.}
    We reduce rich, multidimensional identities to binary contrasts (\leftq{} vs.\ \rightq{} political lean and \aalq{} vs.\ \wmeq{}) and rely on
    imperfect proxies for each: AllSides' editorial lean labels, an LLM
    classifier for \redditpol{} ideology, and the demographic-alignment
    classifier of \citet{blodgett-demographic} for dialect. These axes are coarse, collapsing centrists and ignoring intersectional or non-binary identities, and the labels carry their own error and potential bias. Our
    results should be read as evidence that retrievers are sensitive to these
    signals, not as a complete account of identity-conditioned retrieval.

    \paragraph{Synthetic queries and the paired/unpaired trade-off.}
    Our controlled contrasts use synthetic queries (LLM-generated political queries and morphosyntactically and orthographically augmented \aalq{}), which
    may not faithfully represent the distribution of real user language.
   To mitigate this potential confound, we use approaches that have been human-validated and that aim to avoid substantively altering the meaning of a given text. Therefore, synthetic artifacts may inflate measured biases, but are unlikely to meaningfully drive observed gaps.
    Our quality check
    (\cref{app:paraphrase-qc}) finds recognizable \aalq{} features alongside
    generator artifacts.

    We therefore complement each synthetic set with
    naturally occurring queries, but those are unpaired and confound dialect or
    lean with content and topic.
    Neither setting alone is decisive; we rely on
    their agreement.

    Any systematic stylistic asymmetry the generator introduces between leans (register, presupposition density) is held within---not removed by---the frame control. The \redditpol{} agreement partially mitigates this but does not rule it out.

    \paragraph{Gold-document evaluation.}
    Our health metric scores the rank of each question's original doctor
    response and so measures retrieval competence conditioned on dialect. It does not assess the medical usefulness or safety of the other retrieved
    documents. LLM-judging full retrieved sets along these dimensions is a
    complementary evaluation we leave to future work.

\section*{Ethical considerations}

    \paragraph{What counts as harm.}
    Tailoring retrieval to a user is not inherently harmful, and for health,
    a user's background can legitimately call for different documents (e.g.,
    conditions such as sickle cell disease that differ in prevalence across
    groups). The harms we identify are more specific: in the political domain,
    surfacing documents that confirm a questioner's existing lean can entrench
    echo chambers and contribute to polarization; in the health domain, \aalq{}
    phrasing should not \emph{lower the quality} of the answers a user
    receives. Our concern is thus systematic confirmation and quality
    degradation conditioned on identity, not personalization per se.

    \paragraph{Synthetic \aalq{}.}
    Generating synthetic \aalq{} risks caricaturing the variety, and our
    augmentation pipeline introduces artifacts (\cref{app:paraphrase-qc}). We
    use it only as a controlled diagnostic, pair it with naturally occurring
    \aalq{}, follow pipelines validated with \aalq{} speakers in prior work,
    and do not present it as an authentic sample of the variety.

    \paragraph{Data and privacy.}
    The \redditpol{} ideology labels are inferred by classifiers, not
    self-reported, and we use them only in aggregate. To protect user privacy
 our code release contains Reddit post IDs and labels rather than post text, so that
    content a user later deletes is not redistributed.
    We likewise withhold the per-post classifier rationales, which
    quote the posts, and the ideology-classification code that consumes post
    text; the released per-post query embeddings are dense encodings of the
    withheld titles, so what we withhold is the text itself rather than every
    derivative of it. Every result we report is reproducible from the released
    identifiers and labels, and the post text can be re-fetched by ID for any
    post whose author has not since removed it.

    \paragraph{Potential for misuse.}
    The analysis that diagnoses identity sensitivity (including the linear directions our probes recover) could also be used to \emph{amplify} bias,
    for instance to deliberately serve more confirmatory or lower-quality
    results to a group. We release these resources to support measurement and
    mitigation, and caution against deployments that personalize retrieval
    along identity lines without safeguards.

    \paragraph{Artifacts.} We use artifacts that are publicly available and consistent with their intended use for research.
    \paragraph{AI Use.} We used AI to help with coding and writing. All AI outputs were manually verified.

\section*{Acknowledgements}

    This work was supported in part by National Science Foundation Graduate Research Fellowship DGE-2036197, the Columbia University Provost Diversity Fellowship, and the Columbia School of Engineering and Applied Sciences Presidential Fellowship. Any opinion, findings, and conclusions or recommendations expressed in this material are those of the authors and do not necessarily reflect the views of the National Science Foundation. We thank the anonymous reviewers for discussions and feedback on earlier iterations of this work.

\bibliography{custom}

\appendix
\crefalias{section}{appendix}
\crefalias{subsection}{appendix}
\crefalias{subsubsection}{appendix}

\begin{table*}\centering\small
    \setlength{\tabcolsep}{8pt}
    \begin{tabular}{ll r rrrr}
        \toprule
        & & & \multicolumn{4}{c}{Query length (characters)} \\

        \cmidrule(lr){4-7}
        Dataset & Group & $n$ & Median & Mean & Min & Max \\
        \midrule
        \multirow{2}{*}{\allsidesq{}} & left  & 3{,}723  & 138 & 139 & 98  & 187 \\
                                      & right & 3{,}723  & 139 & 139 & 87  & 192 \\
        \midrule
        \multirow{2}{*}{\redditpol{}} & liberal      & 1{,}327 & 152 & 166 & 98 & 302 \\
                                      & conservative & 1{,}225 & 154 & 168 & 97 & 302 \\
        \midrule
        \multirow{2}{*}{\hcmsynth{}}  & WME & 4{,}999  & 362 & 437 & 63 & 5{,}141 \\
                                      & AAL & 14{,}997 & 384 & 462 & 63 & 5{,}332 \\
        \midrule
        \multirow{2}{*}{\hcmnat{}}    & WME & 659 & 358 & 435 & 119 & 2{,}060 \\
                                      & AAL & 647 & 309 & 352 & 27  & 1{,}724 \\
        \bottomrule
    \end{tabular}
    \caption{Per-group query counts and query-length statistics
    (characters) for the four query datasets. Counts match the evaluated
    sets used throughout: \hcmsynth{} drops one WME source whose AAL
    augmentation was empty ($4{,}999$ WME), and \hcmnat{} applies a
    $\geq\!20$-character noise filter.}
    \label{tab:data-stats}
    \end{table*}
\section{Query dataset statistics}
\label{app:data-stats}

    \cref{tab:data-stats} reports per-group query counts and
    query-length distributions (in characters) for the four query
    datasets. \allsidesq{} and \hcmsynth{} are paired (left/right and
    WME/AAL share content); \redditpol{} and \hcmnat{} are unpaired
    naturalistic pools. \hcmsynth{} produces three AAL paraphrases per WME
    source ($4{,}999\!\times\!3 = 14{,}997$).

\section{Query-side pairing check}
\label{app:pairing-check}

    The \allsidesq{} pipeline (\cref{sec:political-data}) pairs L
    and R queries that share both a source article and a frame, on the
    design intent that frame-level pairing conditions out topical drift
    between partisan queries. We verify this intent on the
    query-embedding side. For each L/R query pair $p =
    (\text{article}, \text{frame})$ and each dense retriever we compute
    $\cos(L_p, R_p)$ under three pairing schemes:
    \emph{same article, same frame} (the \allsidesq{} unit of
    analysis), \emph{same article, different frame} (L from pair $p$,
    R averaged over $5$ random L/R pairs from the same article under
    different frames), and \emph{different article} (R averaged over
    $5$ random pairs from different articles). One of the $1{,}000$
    sampled articles yields no query pairs; of the remaining $999$, $26$
    have a single frame, so the same-article-different-frame contrast
    uses $3{,}697$ paired anchors out of $3{,}723$ total query pairs.
    BM25 is sparse and is omitted.

    \begin{figure}[t]
        \centering
        \includegraphics[width=\linewidth]{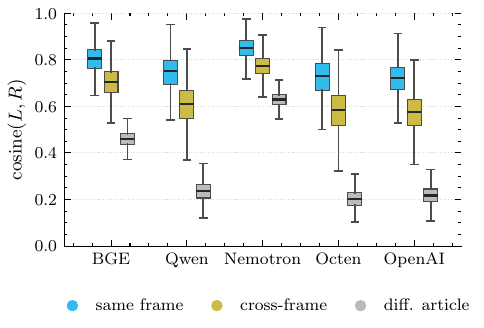}
        \caption{Per-anchor L--R query cosine under three pairing
            schemes (cyan = same article and frame, yellow = same article
            different frame, grey = different article), grouped by
            retriever. Box = IQR with median; whiskers
            $= 1.5\!\times\!$IQR; outliers suppressed. Inter-quartile
            boxes are cleanly separated within every retriever; whiskers
            cross substantially, so individual query pairs overlap
            across schemes even though the means do not.}
        \label{fig:pairing-check}
    \end{figure}

    \cref{fig:pairing-check} shows the per-anchor cosine
    distributions. Mean cosines [95\% cluster-bootstrap CI on the mean,
    $B = 5{,}000$, cluster = article; all CI half-widths $\le 0.006$]
    are: same article + frame $0.717$--$0.850$, same article different
    frame $0.573$--$0.774$, different article $0.204$--$0.630$. Median
    paired $\Delta(\text{within-frame} - \text{cross-frame})$ ranges
    from $+0.075$ (Nemotron) to $+0.139$ (Octen) with cluster-bootstrap
    95\% CIs nowhere crossing zero; paired Wilcoxon signed-rank yields
    $p < 10^{-300}$ on every retriever ($n \approx 3{,}700$).
    Distribution overlap at the per-pair level is non-trivial: whiskers
    ($1.5\!\times\!$IQR) cross substantially on every retriever, so
    individual pairs can have lower same-frame cosine than
    cross-frame cosine. The inter-quartile boxes, however, are cleanly
    separated, and the means are separated by many CI
    half-widths---so the Stage-A/Stage-B \allsidesq{} pipeline behaves as designed at
    the query-embedding level: pairing L and R queries from the same
    source article and the same frame does produce strictly tighter
    topical anchoring than pairing on the article alone.

\section{Labeling Political Frames}
\label{app:political-frames}

    \paragraph{The Boydstun MFC frame set.}
    The Media Frames Corpus (MFC) frame typology
    \cite{boydstun-mfc} defines 15 policy-relevant frames intended to be
    issue-agnostic. They are: \emph{Economic};
    \emph{Capacity \& resources}; \emph{Morality}; \emph{Fairness \&
    equality}; \emph{Legality, constitutionality \& jurisprudence};
    \emph{Policy prescription \& evaluation}; \emph{Crime \&
    punishment}; \emph{Security \& defense}; \emph{Health \& safety};
    \emph{Quality of life}; \emph{Cultural identity}; \emph{Public
    sentiment}; \emph{Political}; \emph{External regulation \&
    reputation}; and \emph{Other}. Each frame is meant to capture a
    distinct ``angle'' on a news story (e.g.\ the same immigration
    article can be framed primarily through \emph{Security} or through
    \emph{Fairness}).

    \paragraph{Stage-A extraction (\allsidesq{}).}
    We prompt \texttt{gpt-5-mini} with the Boydstun definitions and the
    article body, and ask the model to label every frame present with a
    salience score in $[0, 1]$ under a constrained JSON schema (default
    temperature, \texttt{medium} reasoning effort). For each article we
    keep the frames with non-zero salience and feed each
    (article, frame) pair to Stage-B query generation. Across the
    $1{,}000$ sampled articles we extract a mean of $3.72$ frames per
    article.

    \paragraph{Stage-B query prompt.}
    The system prompt asks the model to write three search queries about
    the article, all emphasising the same Boydstun frame: neutral,
    left-leaning, and right-leaning (the neutral variant is unused in this
    paper's left/right contrasts). Hard constraints: same sub-topic across
    the queries; each query clearly emphasises the named frame;
    each query is a single natural search-engine sentence (10--25 words);
    no quotation marks, bullet points, or preambles. The full prompts
    for both stages are shown in \cref{fig:stage-prompts}.

    \begin{figure*}[t]
        \centering
        \begin{minipage}[t]{0.48\textwidth}
            \centering \textbf{Stage A: frame extraction} \\[2pt]
            \begin{tcolorbox}[colback=gray!20, colframe=black]
                \scriptsize
                \textbf{System: } You are an expert media-frame coder using Boydstun et al.'s 15 generic policy frames. \\[2pt]
                The 15 frames (use these keys verbatim): \\
                $\langle$15 ``key --- one-sentence definition'' lines$\rangle$ \\[2pt]
                Read the article and identify every frame that is genuinely present. Return as many or as few frames as the article warrants---do not pad, do not under-report. Rank by salience (most prominent first). Salience is a number in [0, 1]. Use only the keys from the list. Do not invent new frame names. Return strictly valid JSON matching the required schema. \\[1pt]
                \hrule
                \vspace{5pt}
                \textbf{User: } \\[1pt]
                \texttt{Article (article\_id=$\langle$id$\rangle$):} \\
                $\langle$article text$\rangle$
            \end{tcolorbox}
        \end{minipage}\hfill
        \begin{minipage}[t]{0.48\textwidth}
            \centering \textbf{Stage B: query generation} \\[2pt]
            \begin{tcolorbox}[colback=gray!20, colframe=black]
                \scriptsize
                \textbf{System: } You will write three search queries about a news article, all emphasizing the same Boydstun frame: ``$\langle$frame key$\rangle$'' --- $\langle$frame definition$\rangle$ \\[2pt]
                Produce: \\
                -- ``neutral'': the query a non-partisan reader concerned with this frame would type \\
                -- ``left'': the query a progressive/left-leaning reader concerned with this frame would type \\
                -- ``right'': the query a conservative/right-leaning reader concerned with this frame would type \\[2pt]
                Hard constraints: \\
                1. All three queries must concern THE SAME sub-topic of the article. They differ in stance/framing language, not in topic. \\
                2. Each query must clearly emphasize the named frame. \\
                3. Stay grounded in the article's facts; do not invent claims. \\
                4. Each query is a single natural search-engine sentence (10-25 words). \\
                5. No quotation marks, no bullet points, no preambles. \\
                Return strictly valid JSON matching the required schema. \\[1pt]
                \hrule
                \vspace{5pt}
                \textbf{User: } \\[1pt]
                \texttt{Article (article\_id=$\langle$id$\rangle$):} \\
                $\langle$article text$\rangle$ \\[2pt]
                \texttt{Frame to emphasize: $\langle$frame key$\rangle$ --- $\langle$frame definition$\rangle$}
            \end{tcolorbox}
        \end{minipage}
        \caption{Stage-A (frame extraction) and Stage-B (frame-conditioned
        query generation) prompts (\texttt{gpt-5-mini}, \texttt{medium}
        reasoning effort, default temperature). $\langle\cdot\rangle$ marks
        per-request substitutions: Stage A receives the 15
        ``key---definition'' lines for the frames listed above; Stage B is
        instantiated once per (article, frame) cell. Both stages decode
        under strict JSON schemas---Stage A returns
        \texttt{\{frames: [\{key, salience\}]\}}, Stage B returns
        \texttt{\{neutral, left, right\}}. The neutral query is generated
        but unused in this paper's left/right contrasts.}
        \label{fig:stage-prompts}
    \end{figure*}

\section{Reddit Data Collection}
\label{app:reddit-collection}

    We scrape the most recent $15,000$ posts from each subreddit as candidate questions.

    \paragraph{Ideology Labeling. }
        For each post, we take the post text (excluding the question) and provide it two LLMs, prompting them to identify the most likely ideology of the author (i.e., liberal, centrist, or conservative). Specifically, we prompt \texttt{gpt-5.4-mini} and Qwen3.5-4B (\href{https://huggingface.co/Qwen/Qwen3.5-4B}{\texttt{Qwen/Qwen3.5-4B}}; \citealp{qwen35blog}). Notably, the title question used in embedding evaluations is not provided to the model given that the posts often contain more context specific to the author and to avoid potential data leakage; for example, a Qwen model that predicts a question is conservative-leaning may indicate that the associated embedding model may be predisposed to embed similarly to conservative-leaning texts. The provided prompt is shown in \cref{fig:ideology-prompt}. We extract only posts for which both models agreed on the most likely label to form the final dataset of queries. Following the privacy commitment in our Ethical considerations, the code release contains the resulting post IDs and labels but not the post text, the per-post classifier rationales, or the classification code itself.

    \begin{figure}[htbp]
        \centering
        \begin{minipage}[t]{0.95\columnwidth}
            \begin{tcolorbox}[colback=gray!20, colframe=black]
                \scriptsize
                \textbf{System: } You are an expert in analysis of political rhetoric. \\[1pt]
                \hrule
                \vspace{5pt}
                \textbf{User: } Determine whether the given message is most likely written by a liberal, conservative, or centrist author. If there is no evidence whatsoever of the author's political leaning, output "not evident", although this should only be used in cases where there is truly no most likely option. The last word of your response should strictly be one of the possible labels ("liberal", "conservative", "centrist", "not evident") in double quotes with nothing after and no qualifiers. \\ Message: "{\color{blue} This not my perspective whatsoever, but I see \textbf{conservatives} all over the place cheering what happened as an example of a president \textbf{"finally standing up for America."} They're saying that POTUS' behaviors were a sign of patriotic strength, as opposed to the "dodged bullet" of a possible Harris/Walz administration's milquetoast capitulation to Ukraine. \textbf{Conservatives really do see what happened as American flexing its muscle and showing strength, and they're proud of it.}\\How should \textbf{we} refute of this point of view?}"\\[1pt]
                \hrule
                \vspace{5pt}
                \textbf{Assistant: } {\color{darkgreen} This message is most likely written by a "liberal"} \\
            \end{tcolorbox}
        \end{minipage}
        \caption{Prompt used for ideology classification. {\color{blue} Blue text} represents an example post, while {\color{darkgreen} green text} represents the associated model prediction. Particular phrases that may signal the political ideology of the author are \textbf{bolded}.}
        \label{fig:ideology-prompt}
    \end{figure}

\section{Model Details}
\label{app:model-details}

    \cref{tab:model-dets} lists the evaluated retrieval models: four
    open-weight dense encoders run through \texttt{sentence-transformers},
    the closed \teLargeShort{} accessed through the OpenAI API, and the
    BM25-Okapi sparse baseline run through \texttt{bm25s} \cite{bm25s}.

    \begin{table}
        \centering
        \resizebox{.95\linewidth}{!}{
            \begin{tabular}{llll}
                \toprule
                Model & Size & Emb.\ dim & Checkpoint \\
                \midrule
                BGE-large          & 0.3B & 1024 & \href{https://huggingface.co/BAAI/bge-large-en-v1.5}{\texttt{BAAI/bge-large-en-v1.5}} \\
                Qwen3-Embedding-8B & 8B   & 4096 & \href{https://huggingface.co/Qwen/Qwen3-Embedding-8B}{\texttt{Qwen/Qwen3-Embedding-8B}} \\
                Llama-Nemotron-8B  & 8B   & 4096 & \href{https://huggingface.co/nvidia/llama-embed-nemotron-8b}{\texttt{nvidia/llama-embed-nemotron-8b}} \\
                Octen-Embedding-8B & 8B   & 4096 & \href{https://huggingface.co/Octen/Octen-Embedding-8B}{\texttt{Octen/Octen-Embedding-8B}} \\
                \teLargeShort{}    & ---  & 3072 & \texttt{text-embedding-3-large} (API) \\
                BM25-Okapi         & ---  & ---  & \texttt{bm25s} (sparse) \\
                \bottomrule
            \end{tabular}
        }
        \caption{Evaluated retrieval models. \teLargeShort{} uses its
        default $3072$ dimensions (no Matryoshka truncation); BM25 is
        sparse/lexical and has no dense embedding.}
        \label{tab:model-dets}
    \end{table}

    \paragraph{Implementation.}
    All embeddings are L2-normalised, so retrieval ranks by cosine
    similarity, and inputs are capped at $512$ tokens. The three 8B
    encoders run in \texttt{bfloat16} and BGE-large in \texttt{float32};
    \teLargeShort{} returns $3072$-dimensional vectors via the API.
    BM25 (Okapi) is computed with \texttt{bm25s} \cite{bm25s} for all
    corpora, roughly $200\times$ faster than \texttt{rank\_bm25} at the
    $112$k-document scale.
    Each encoder uses its documented query/document convention:
    \begin{itemize}
        \setlength{\itemsep}{0pt}
        \item \textbf{BGE-large}: queries are prefixed with
        ``\texttt{Represent this sentence for searching relevant
        passages:\ }''; documents take no prefix.
        \item \textbf{Qwen3-Embedding-8B}: the model's built-in
        \texttt{query}/\texttt{document} prompt templates.
        \item \textbf{Llama-Nemotron-8B}: queries are prefixed with an
        \texttt{Instruct: \ldots\textbackslash nQuery:\ } retrieval
        instruction; documents take no prefix.
        \item \textbf{Octen-Embedding-8B}: queries take no prefix;
        documents are prefixed with ``\texttt{-\ }'' (the model card's
        recommended workaround).
        \item \textbf{\teLargeShort{}}: no query/document distinction.
    \end{itemize}

\section{\redditpol{} LLM-judged gold-filter}
\label{app:reddit-judge}

    \begin{table*}[t]\centering\small
    \begin{tabular}{lrrrr}
        \toprule
         & raw & gpt-only & Qwen-only & \textbf{gold} \\
        Retriever & R$-$L & R$-$L & R$-$L & \textbf{R$-$L ($p$)} \\
        \midrule
        BGE-large       & $+0.020$ & $+0.069$ & $+0.088$ & $+0.109$ (\textbf{$0.007$}) \\
        Qwen3-Emb-8B    & $+0.025$ & $+0.065$ & $+0.080$ & $+0.069$ ($0.047$) \\
        Llama-Nemotron  & $+0.033$ & $+0.072$ & $+0.174$ & $+0.186$ (\textbf{$10^{-3}$}) \\
        Octen-Emb-8B    & $+0.048$ & $+0.043$ & $+0.087$ & $+0.104$ (\textbf{$0.002$}) \\
        \teLargeShort{} & $+0.056$ & $+0.102$ & $+0.107$ & $+0.136$ (\textbf{$10^{-4}$}) \\
        BM25            & $+0.009$ & $+0.151$ & $+0.169$ & $\mathbf{+0.195}$ (\textbf{$0.006$}) \\
        \bottomrule
    \end{tabular}
    \caption{Reddit right$-$left mean lean@$10$ gap under four filter
    definitions: raw (no filter, $n=2{,}552$ queries per encoder);
    gpt-5-mini-only; Qwen-only; and gold
    (both judges grade the article relevant; used in the main figure).
    All four agree on
    direction; the tighter the filter, the larger the gap. The
    gold column is the primary contrast used in
    \cref{fig:political-combined}b.}
    \label{tab:reddit-judge}
    \end{table*}

    \paragraph{Setup.}
    For \redditpol{} there is no editorially-aligned
    story-counterpart gold the way the AllSides roundup provides one
    for synthetic query pairs. We use an LLM-judge gold from two
    independent relevance judges, both run on every
    $(\text{query}, \text{article})$ pair in the top-10 retrieval
    pool: \textbf{gpt-5-mini} (closed-weights, OpenAI batch API) and
    \textbf{Qwen3.5-35B-A3B} (open-weights, 35B-total / 3B-active
    sparse MoE; run locally via vLLM with PP=4 on $4\times$ RTX
    4090). Both use the same binary $0/1$ schema with the same
    prompt and $T{=}0$. All $2{,}559$ left + right queries are judged on
    their top-10 retrievals across the $5$ dense retrievers + BM25,
    deduped to $\mathbf{102{,}637}$ unique pairs.

    \paragraph{Inter-judge agreement.}
    Cohen's $\kappa = 0.50$; raw pairwise agreement $89.2\%$.
    gpt-5-mini is the more permissive judge ($15.4\%$ relevance
    rate vs.\ $9.2\%$ for Qwen). Qwen's ``relevant'' set is
    approximately a subset of gpt's: gpt-5-mini recall vs.\
    Qwen-as-gold is $74.9\%$, while Qwen's recall vs.\
    gpt-as-gold is only $44.6\%$—agreement is asymmetric and
    consistent with a calibration difference, not directional
    disagreement.
    Per-side $\kappa$: left $=0.500$, right $=0.502$
—\emph{judges do not disagree differently by query side},
    which is the key robustness check against the worry that the
    closed-weights judge might encode ideologically biased
    relevance grades.

    \paragraph{Gold filter.}
    A retrieval is ``gold'' for the primary test
    (\cref{fig:political-combined}b) iff \emph{both} judges
    grade it $1$. This passes $7{,}047$ pairs ($6.87\%$ of the
    pool). Per-encoder retention varies: \teLargeShort{} $14.3\%$,
    Qwen3-Emb-8B $14.0\%$, Octen $12.4\%$, BGE $10.7\%$,
    Llama-Nemotron-8B $5.9\%$, BM25 $3.5\%$. The two encoders with
    the lowest retention (Nemotron, BM25) still surface enough
    relevant docs to produce significant gaps, and BM25 in fact
    carries the \emph{largest} right$-$left gap of any retriever
    under this gold. \cref{tab:reddit-judge} reports the
    per-retriever right$-$left gap under each filter definition
    (raw, each judge alone, and gold).

    \paragraph{Judge prompt.}
    Both judges see the binary relevance prompt in
    \cref{fig:relevance-judge-prompt} at $T{=}0$. The \allsidesq{}
    LLM-judge ablation (\cref{app:gold-filter}) uses the same template
    and binary rubric with news-query wording; both prompts are shown.

    \begin{figure*}[t]
        \centering
        \begin{minipage}[t]{0.48\textwidth}
            \centering \textbf{\redditpol{}} \\[2pt]
            \begin{tcolorbox}[colback=gray!20, colframe=black]
                \scriptsize
                \textbf{System: } You are a careful retrieval relevance judge. You see a question from a Reddit user and a news article that a retriever returned for it. Decide whether the article would help that user reach an answer. Return only valid JSON matching the required schema; no extra text. \\[1pt]
                \hrule
                \vspace{5pt}
                \textbf{User: } \\[1pt]
                \texttt{Question (Reddit post):} \\
                $\langle$query text$\rangle$ \\[2pt]
                \texttt{Retrieved article:} \\
                $\langle$article text$\rangle$ \\[2pt]
                Does the article help the Reddit user reach an answer to their question? Grade 1 if on-topic and at least partly informative; 0 otherwise. Return JSON: \texttt{\{"grade": 0 or 1\}}.
            \end{tcolorbox}
        \end{minipage}\hfill
        \begin{minipage}[t]{0.48\textwidth}
            \centering \textbf{\allsidesq{}} \\[2pt]
            \begin{tcolorbox}[colback=gray!20, colframe=black]
                \scriptsize
                \textbf{System: } You are a careful retrieval relevance judge. You see a search query about a news story and a news article that a retriever returned for it. Decide whether the article is on-topic for the query---about the same news story or issue. Return only valid JSON matching the required schema; no extra text. \\[1pt]
                \hrule
                \vspace{5pt}
                \textbf{User: } \\[1pt]
                \texttt{Search query:} \\
                $\langle$query text$\rangle$ \\[2pt]
                \texttt{Retrieved article:} \\
                $\langle$article text$\rangle$ \\[2pt]
                Is the article on-topic for this query---about the same news story or issue? Grade 1 if on-topic; 0 otherwise. Return JSON: \texttt{\{"grade": 0 or 1\}}.
            \end{tcolorbox}
        \end{minipage}
        \caption{Binary relevance-judge prompts, run by both judges
        (\texttt{gpt-5-mini} and \texttt{Qwen3.5-35B-A3B}, $T{=}0$): the
        \redditpol{} two-judge gold (\cref{app:reddit-judge}) and the
        \allsidesq{} LLM-judge ablation (\cref{app:gold-filter}). Both
        share the binary $0/1$ rubric and JSON \texttt{\{"grade"\}} schema;
        only the domain wording differs---\redditpol{} asks whether the
        article helps the user \emph{answer} their question, \allsidesq{}
        whether it is \emph{on-topic} for the news story.}
        \label{fig:relevance-judge-prompt}
    \end{figure*}

\section{Gold-filtering robustness}
\label{app:gold-filter}

    Our political results measure retrieval lean over the documents a
    retriever surfaces that are relevant to the query
    (\cref{sec:political-methods}). ``Relevant'' is defined
    differently on the two datasets---same editorial roundup on
    \allsidesq{}, two-judge LLM relevance on \redditpol{}---so we verify
    that the own-lean gap does not depend on that choice. We examine four
    settings; the gap is present and significant in three, and
    the diagnostics below show the lone exception is a property of
    \redditpol{} retrieval, not of the metric.

\begin{figure*}[t]
        \centering
        \includegraphics[width=\linewidth]{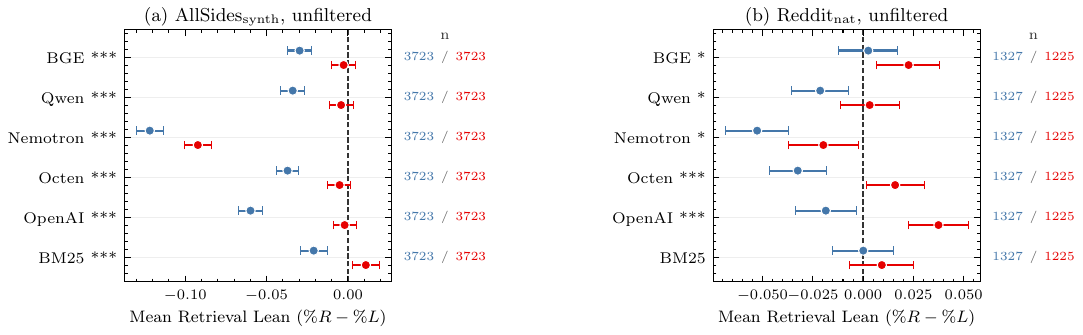}
        \caption{Per-retriever mean retrieval lean over the \emph{full}
        top-10 (no relevance filter), 95\% bootstrap CIs.
        \textbf{(a)} \allsidesq{} \leftq{}/\rightq{} queries;
        \textbf{(b)} \redditpol{} liberal-/conservative-asker queries.
        Stars on retriever labels (\texttt{*}$p{<}0.05$,
        \texttt{**}$p{<}0.01$, \texttt{***}$p{<}0.001$); $n$ per group in
        the right margin. The own-lean gap (\rightq{} marker right of
        \leftq{}) holds without any filtering.}
        \label{fig:gold-unfiltered}
    \end{figure*}

    \paragraph{The gap appears with no relevance filter.}
    Taking the mean lean over the \emph{full} top-10 of every query, with
    no relevance filter at all, the own-lean gap is already present
    (\cref{fig:gold-unfiltered}): every \allsidesq{} retriever is
    significant ($p<10^{-8}$), as are five of six on \redditpol{}
    ($p<0.05$; only BM25, whose raw retrievals carry no lean, is not).
    The filter sharpens the signal rather than creating it.

\begin{figure}[t]
        \centering
        \includegraphics[width=0.92\linewidth]{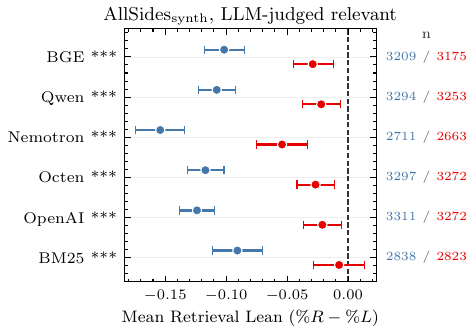}
        \caption{\allsidesq{} per-retriever lean over LLM-judged-relevant
        top-10 docs (the \redditpol{} setting applied to
        \allsidesq{}): the own-lean gap holds on all six retrievers
        ($p<10^{-10}$), larger than under the editorial gold
        (\cref{tab:political-goldtest}). 95\% bootstrap CIs; $n$ =
        retained left/right queries.}
        \label{fig:gold-synth-llm}
    \end{figure}

\begin{figure}[t]
        \centering
        \includegraphics[width=0.92\linewidth]{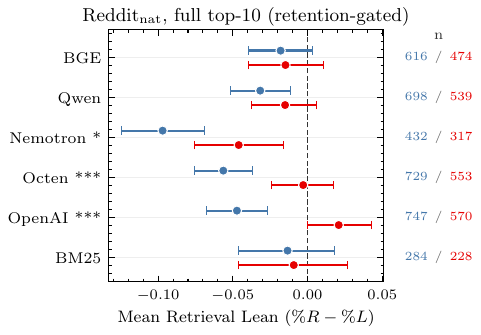}
        \caption{\redditpol{} per-retriever lean over the \emph{full}
        top-10, restricted to queries with $\geq 1$ relevant doc (the
        \allsidesq{} setting applied to \redditpol{}). Only
        three of six retrievers stay significant---the off-topic majority
        of the top-10 dilutes the signal (\cref{fig:gold-reldist}).}
        \label{fig:gold-reddit-full10}
    \end{figure}

    \paragraph{The two relevance methods are interchangeable.}
    The primary \redditpol{} metric restricts lean to LLM-judged-relevant
    documents (\cref{app:reddit-judge}). Applying the \emph{same}
    two-judge relevance gold to \allsidesq{} (same prompt,
    \cref{fig:relevance-judge-prompt})---an independent judgment on
    all $242{,}000$ \leftq{}/\rightq{} top-10 pairs, gold iff both
    \texttt{gpt-5-mini} and \texttt{Qwen3.5-35B-A3B} grade
    relevant---reproduces the own-lean gap on all six retrievers, larger
    than the editorial-gold gap ($+0.06$ to $+0.09$ vs.\ $+0.02$ to
    $+0.05$; \cref{fig:gold-synth-llm}), since the relevant subset
    concentrates the signal. Conversely, applying the \allsidesq{}-style
    full-top-10 mean to \redditpol{} weakens it: only three of six
    retrievers stay significant (\cref{fig:gold-reddit-full10}).

\begin{figure}[t]
        \centering
        \includegraphics[width=0.82\linewidth]{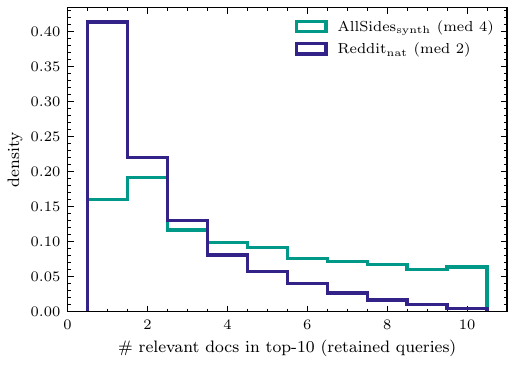}
        \caption{Number of relevant docs in the top-10 per
        (query, retriever), among queries with $\geq 1$ relevant.
        \allsidesq{} lists are far more on-topic than \redditpol{}'s
        ($83\%$ vs.\ $40\%$ of lists contain any relevant doc).}
        \label{fig:gold-reldist}
    \end{figure}

    \paragraph{Why the full-top-10 \redditpol{} gap is diluted, not absent.}
    The datasets differ sharply in how much of the top-10 is on-topic
    (\cref{fig:gold-reldist}). On \allsidesq{}, $83\%$ of
    (query, retriever) top-10 lists contain $\geq 1$ relevant article
    (median $4$ among those), because the synthetic queries are written
    from the corpus; on \redditpol{}, only $40\%$ do (median $2$). A
    full-top-10 mean on \redditpol{} therefore averages over a large
    majority of off-topic articles---which carry no consistent lean---so
    the relevant-subset metric is the appropriate one there.
    BM25 is the extreme case of this dilution: its gold retention is the lowest of any retriever (3.5\%; \cref{app:reddit-judge}), so its full top-10 carries no consistent lean (\cref{fig:gold-unfiltered}b) even though its gap over relevant documents is the largest of any retriever (\cref{tab:reddit-judge}). The converse holds on \allsidesq{}, where queries are written from the corpus and overlap is high; even there, shared topical vocabulary (e.g., "CBO", "\$10.10"; \cref{tab:data-exs}) dominates the BM25 score and matches articles of both leans within a roundup, leaving the few partisan terms to shift the top-10 only modestly.

\begin{figure}[t]
        \centering
        \includegraphics[width=0.55\linewidth]{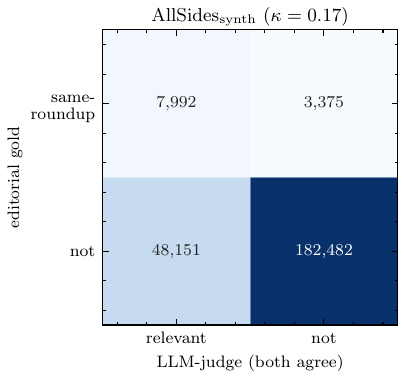}
        \caption{\allsidesq{} LLM-judge relevance (both judges agree) vs.\
        editorial same-roundup gold, over all $242{,}000$ judged top-10
        pairs. The judge is broader than the editorial set ($\kappa=0.17$),
        yet the own-lean gap holds under both.}
        \label{fig:gold-confusion}
    \end{figure}

    \paragraph{Robustness to the relevance judge.}
    On \allsidesq{} we have editorial ground truth (same-roundup
    membership), letting us measure how far the LLM judge departs from it
    (\cref{fig:gold-confusion}). The two-judge gold is much broader
    than the editorial set: it recovers $70\%$ of same-roundup articles
    but also flags many other on-topic ones, for low agreement
    (Cohen's $\kappa = 0.17$). Despite this, the own-lean gap is
    significant under \emph{both} definitions (editorial:
    \cref{tab:political-goldtest}; LLM-judge:
    \cref{fig:gold-synth-llm}), so it does not hinge on the exact
    relevance criterion.

\section{Significance tests}
\label{app:sigtests}

    The own-lean and dialect contrasts are tested as follows; per-retriever
    $p$-values are in \cref{app:political-goldtest} (political) and
    \cref{app:hcmagic-real} (dialect).

    \paragraph{Paired contrasts (\allsidesq{}, \hcmsynth{}, translation).}
    On the paired datasets we test the mean per-pair difference
    ($\dlean$ or $\drank$) against a null of no preference in two ways. The
    \emph{sign-flip permutation} \citep{pitman-permutation} randomly flips
    the sign of each per-pair difference on $B{=}50{,}000$ resamples and
    asks how often the permuted mean is at least as extreme as the observed
    mean (two-sided; $p$-floor $1/B = 2\!\cdot\!10^{-5}$). The two-sided
    \emph{paired Wilcoxon signed-rank} test \citep{wilcoxon-signedrank}
    asks whether the per-pair difference has a non-zero median across pairs,
    without distributional assumptions.

    \paragraph{Unpaired contrast (\redditpol{}, \hcmnat{}).}
    The naturalistic datasets are unpaired (different queries in each
    group), so we use a two-sided rank-based \emph{Mann--Whitney $U$} test
    \citep{mann-whitney}, asking whether one group's per-query lean (or
    MRR) is stochastically larger than the other's.

\section{Per-retriever gold-filtered own-lean tendency}
\label{app:political-goldtest}

    On \allsidesq{}, the per-retriever paired $\overline{\dlean} = \lean(R) - \lean(L)$ at
    $k{=}10$ on the fair corpus, restricted to (article, frame) pairs
    where both the left and the right query surface a same-story
    counterpart in top-10 (the gold set, \cref{tab:political-goldtest}).
    This is the per-retriever difference behind
    \cref{fig:political-combined}a, which plots the left- and
    right-query lean separately. Sign-flip permutation ($B{=}50{,}000$,
    $p$-floor $2\!\cdot\!10^{-5}$) and paired Wilcoxon signed-rank both
    reject on every retriever.

    \begin{table*}[!t]\centering\small
    \setlength{\tabcolsep}{10pt}
    \begin{tabular}{lrrrrr}
        \toprule
        Retriever & $n$ pairs & $\overline{\dlean}$ (R$-$L) & 95\% CI & Wilcoxon $p$ & Perm $p$ \\
        \midrule
        BGE-large            & 2{,}422 & $+0.027$ & $[+0.016, +0.038]$ & $4.4\!\cdot\!10^{-6}$ & $<2\!\cdot\!10^{-5}$ \\
        Qwen3-Emb-8B         & 2{,}508 & $+0.023$ & $[+0.014, +0.033]$ & $2.0\!\cdot\!10^{-6}$ & $<2\!\cdot\!10^{-5}$ \\
        Llama-Nemotron-8B    & 1{,}416 & $+0.025$ & $[+0.011, +0.038]$ & $9.5\!\cdot\!10^{-5}$ & $2.8\!\cdot\!10^{-4}$ \\
        Octen-Emb-8B         & 2{,}605 & $+0.023$ & $[+0.014, +0.032]$ & $1.0\!\cdot\!10^{-5}$ & $<2\!\cdot\!10^{-5}$ \\
        \teLargeShort{}      & 2{,}663 & $\mathbf{+0.045}$ & $[+0.035, +0.054]$ & $3.9\!\cdot\!10^{-21}$ & $<2\!\cdot\!10^{-5}$ \\
        BM25 (Okapi)         & 1{,}798 & $\mathbf{+0.043}$ & $[+0.029, +0.057]$ & $1.4\!\cdot\!10^{-8}$ & $<2\!\cdot\!10^{-5}$ \\
        \bottomrule
    \end{tabular}
    \caption{\allsidesq{} gold-filtered paired contrast at $k{=}10$;
    positive $\overline{\dlean}$ means each query retrieves toward its
    own political lean. Both paired tests reject on every retriever:
    two-sided paired Wilcoxon and sign-flip permutation
    ($B{=}50{,}000$, $p$-floor $2\!\cdot\!10^{-5}$;
    \cref{app:sigtests}).}
    \label{tab:political-goldtest}
    \end{table*}

    Queries on \redditpol{} are unpaired, so the parallel per-retriever
    contrast is the \emph{own-lean gap}: the mean gold-filtered lean over
    conservative-asker queries minus that over liberal-asker queries
    (\cref{tab:reddit-goldtest}), the difference behind
    \cref{fig:political-combined}b. A two-sided Mann--Whitney $U$ test
    rejects on every retriever ($p<0.05$).

    \begin{table*}[!t]\centering\small
    \setlength{\tabcolsep}{10pt}
    \begin{tabular}{lrrrr}
        \toprule
        Retriever & $n$ (lib/con) & gap (con$-$lib) & 95\% CI & MWU $p$ \\
        \midrule
        BGE-large            & $616/474$ & $+0.109$ & $[+0.028, +0.193]$ & $7.1\!\cdot\!10^{-3}$ \\
        Qwen3-Emb-8B         & $698/539$ & $+0.069$ & $[-0.002, +0.139]$ & $4.7\!\cdot\!10^{-2}$ \\
        Llama-Nemotron-8B    & $432/317$ & $\mathbf{+0.186}$ & $[+0.079, +0.297]$ & $1.1\!\cdot\!10^{-3}$ \\
        Octen-Emb-8B         & $729/553$ & $+0.103$ & $[+0.037, +0.176]$ & $1.5\!\cdot\!10^{-3}$ \\
        \teLargeShort{}      & $747/570$ & $+0.136$ & $[+0.064, +0.205]$ & $7.7\!\cdot\!10^{-5}$ \\
        BM25 (Okapi)         & $284/228$ & $\mathbf{+0.195}$ & $[+0.062, +0.323]$ & $5.5\!\cdot\!10^{-3}$ \\
        \bottomrule
    \end{tabular}
    \caption{\redditpol{} gold-filtered own-lean gap at $k{=}10$, parallel
    to \cref{tab:political-goldtest}: mean lean over conservative-asker
    queries minus mean lean over liberal-asker queries, computed over
    LLM-judged-relevant documents. Positive means each asker retrieves
    toward their own political lean. $n$ is the retained liberal/conservative
    query count; 95\% bootstrap CI on the gap; two-sided Mann--Whitney $U$
    $p$ (the test behind \cref{fig:political-combined}b). Qwen3-Emb-8B's
    gap CI marginally includes $0$ although its Mann--Whitney $p<0.05$.}
    \label{tab:reddit-goldtest}
    \end{table*}

\section{Political k-sweep companions}
\label{app:political-ksweeps}

    Companion k-sweep figures to
    \cref{fig:political-combined}. The L--R separation (a) and the
    asker-ideology right$-$left gap (b) are stable across
    $k\in\{5, 10, 20, 50, 100\}$ in both \allsidesq{} and \redditpol{}.

    \begin{figure}[h]
        \centering
        \includegraphics[width=0.95\linewidth]{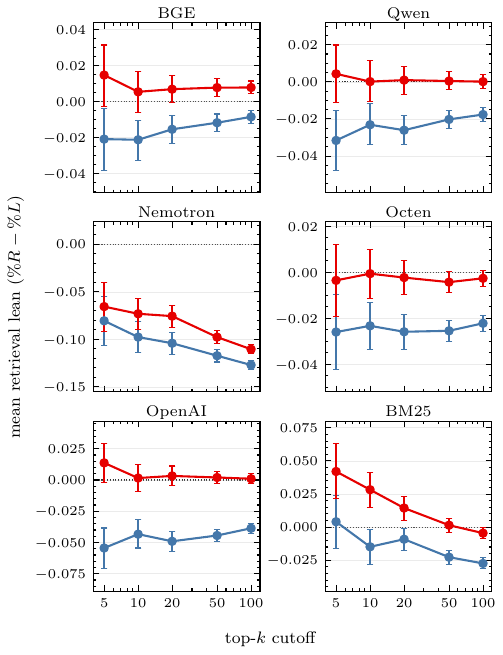}
        \caption{\allsidesq{} gold-filtered paired lean across
        $k\in\{5, 10, 20, 50, 100\}$, per retriever (per-panel $y$-axis).
        Whiskers: 95\% cluster-bootstrap CIs.}
        \label{fig:political-ksweep}
    \end{figure}

    \begin{figure}[h]
        \centering
        \includegraphics[width=0.95\linewidth]{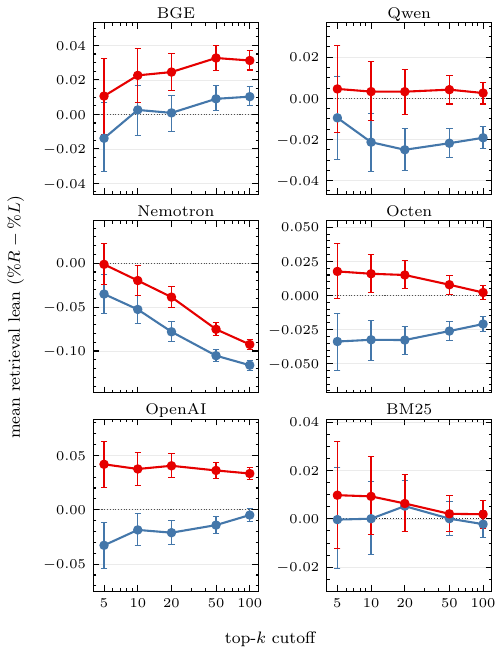}
        \caption{\redditpol{} mean retrieval lean by asker ideology
        across $k\in\{5, 10, 20, 50, 100\}$, per retriever (per-panel
        $y$-axis; raw unfiltered, all retrievals).
        Whiskers: 95\% cluster-bootstrap CIs.}
        \label{fig:political-real-arm-ksweep}
    \end{figure}

\section{HealthCareMagic corpus example}
\label{app:hcmagic-corpus-ex}

    \cref{tab:hcmagic-corpus-ex} gives a worked example from each dataset
    (queries from \cref{tab:aalwme-exs}), with the real doctor response each
    query is scored against---indexed from the originating HealthCareMagic row.
    \hcmsynth{} is paired: a real WME query and its synthetic AAL rewrite share
    one gold document. \hcmnat{} is unpaired, so it contributes a real AAL post
    and a separate real WME post, each a different patient with its own gold.
    All are ranked against the same $112{,}165$-document corpus.

    \begin{table*}[t]\centering\small
    \renewcommand{\arraystretch}{1.3}
    \begin{tabular}{@{}p{0.24\textwidth} p{0.70\textwidth}@{}}
        \toprule
        \multicolumn{2}{@{}l}{\textbf{\hcmsynth{}}\quad---\quad real WME query, synthetic AAL rewrite, shared real gold document} \\
        \midrule
        AAL query \emph{(synthetic)} & \textit{Hi {\color{aalcolor}docta}, how {\color{aalcolor}might can} that renal {\color{aalcolor}functions} be monitored in {\color{aalcolor}one} patient undergoing hemodialysis 3 times per week?} \\
        WME query \emph{(real)} & \textit{Hi doctor, how can the renal function be monitored in a patient undergoing hemodialysis 3 times per week?} \\
        Gold document \emph{(real)} & Renal function is monitored by maintaining a chart of blood urea and serum creatinine levels\ldots during each visit of dialysis you must\ldots note down the\ldots levels of renal parameters\ldots regularly\ldots show it to the nephrologist\ldots Take care. \\
        \midrule
        \multicolumn{2}{@{}l}{\textbf{\hcmnat{}}\quad---\quad real AAL post and a separate real WME post, each with its own gold (unpaired)} \\
        \midrule
        AAL query \emph{(real)} & \textit{\ldots about 2 weeks ago she {\color{aalcolor}notice} she had a lump just below her left rib in the front\ldots once {\color{aalcolor}ahe straight} again u can see it and feel it. What can this be.} \\
        Gold document \emph{(real)} & Hi, Dear, thanks for the query to Chat Doctor\ldots In my opinion your daughter has a third nipple or a Supernumerary nipple below the normal nipple on left side, or may be a puberty change\ldots Have a good time. \\
        \addlinespace
        WME query \emph{(real)} & \textit{Every now and then I will feel my chest tighten and my heart starts pounding extremely hard. I cant breath and I begin feeling very lightheaded. It usually only lasts a few seconds and does not happen too often. Any ideas as to what could be causing this?} \\
        Gold document \emph{(real)} & Hello and thanks for writing\ldots The symptoms you have described are somewhat red flags and need a thorough workup to get to a cause\ldots conditions that might cause them include cardiac arrhythmia, angina, anxiety or Blood Pressure issues\ldots You may also need few tests like BP measurement, CBC, FT, ECG, Holder study\ldots I suggest you consult a physician and get yourself worked up for the problem. \\
        \bottomrule
    \end{tabular}
    \caption{Worked query--document examples for both dialect datasets (the
    queries from \cref{tab:aalwme-exs}), each scored against a \emph{real} gold
    document---the doctor response indexed from its originating HealthCareMagic
    row (abridged with \ldots). \hcmsynth{} is paired: a \emph{real} WME query
    and its \emph{synthetic} AAL rewrite share one gold document. \hcmnat{} is
    unpaired, so we show a \emph{real} AAL post and a separate \emph{real} WME
    post---different patients, each with its own gold document
    (\cref{sec:aalwme-data}). AAL surface markers in
    {\color{aalcolor}purple}.}
    \label{tab:hcmagic-corpus-ex}
    \end{table*}

\section{HCMagic input filter and dropped rows}
\label{app:hcmagic-filter}

    \paragraph{\hcmnat{} (naturalistic).}
    The hcmagic source contains $1{,}315$ rows (AAL: $656$; WME: $659$).
    The loader applies two filters before embedding/retrieval, both
    motivated by inspection of the raw query field:
    (i) drop rows with missing/blank query text (otherwise pandas
    serialises NaN as the literal string \texttt{"nan"} and the embedding
    step proceeds on that junk text);
    (ii) drop rows whose query is shorter than 20 characters after
    stripping whitespace, since the bottom of the AAL length distribution
    is dominated by one-word gibberish/typos
    (e.g.\ ``thtr'', ``aesthema'', ``zzxczxc'') and one three-word
    fragment (``i wana pregnent'') that do not function as questions.
    The 20-char threshold preserves real-but-brief medical questions
    such as ``bipolar type 2 can be cure?'' (27 chars).
    Nine AAL rows are dropped total---one by filter (i) and eight by
    filter (ii)---yielding a final AAL count of $647$. WME has zero
    rows dropped (minimum length $119$ chars). The same nine rows are
    excluded from trans-AAL via the paired-on-row join, so the paired
    contrast runs on $n{=}647$ pairs.

    \paragraph{\hcmsynth{} (synthetic).}
    The \hcmsynth{} source contains $5{,}000$ rows. The same
    20-character minimum filter is applied to all four query variants
    (\texttt{input} plus three \texttt{aal\_message\{1,2,3\}}
    paraphrases) and the \texttt{output} field; one row is dropped
    (all three AAL paraphrase fields empty), yielding $n{=}4{,}999$
    paired samples used in \cref{tab:hcmagic-real} and
    \cref{fig:aalwme-combined}a.

    \begin{table}[h]\centering\small
    \setlength{\tabcolsep}{6pt}
    \begin{tabular}{rrl}
        \toprule
        idx & char len & query (first 30 chars) \\
        \midrule
        70  & 7  & \texttt{zzxczxc} \\
        144 & 9  & \texttt{lipoprint} \\
        267 & 7  & \texttt{nothing} \\
        386 & 4  & \texttt{thtr} \\
        398 & 12 & \texttt{bhjkhjkhklhk} \\
        456 & 8  & \texttt{aesthema} \\
        495 & 15 & \texttt{i wana pregnent} \\
        511 & 0  & \texttt{(empty)} \\
        569 & 4  & \texttt{SDfc} \\
        \bottomrule
    \end{tabular}
    \caption{The nine AAL rows dropped from the hcmagic source by the
    input-validity filter.}
    \label{tab:hcmagic-dropped}
    \end{table}

\section{Per-retriever dialect gap}
\label{app:hcmagic-real}

    The per-retriever contrasts behind \cref{fig:aalwme-combined}, in
    the same format as the political tables
    (\cref{app:political-goldtest}). Positive values mean the WME-form
    query retrieves better---its answer document ranks higher---than the
    AAL form. \hcmsynth{} is \emph{paired}, summarised by the mean per-sample
    $\drank = \mathrm{RR}_{\text{WME}} - \mathrm{RR}_{\text{AAL}}$ over
    reciprocal rank and tested with paired Wilcoxon and sign-flip
    permutation; \hcmnat{} is an \emph{unpaired} between-group MRR gap on
    naturally-occurring queries (different patients), tested with a
    two-sided Mann--Whitney $U$. All at $k{=}10$ on the unified
    $112{,}165$-document corpus, no relevance filter; 95\% CIs are
    bootstrap. No multiple-comparison correction; the BGE \hcmnat{} entry
    ($p{=}0.10$) is the only contrast that does not clear $p<0.05$.

    \begin{table*}[t]\centering\small
    \setlength{\tabcolsep}{10pt}
    \begin{tabular}{lrrrrr}
        \toprule
        Retriever & $n$ pairs & $\overline{\drank}$ (WME$-$AAL) & 95\% CI & Wilcoxon $p$ & Perm $p$ \\
        \midrule
        BGE-large            & 4{,}999 & $\mathbf{+0.050}$ & $[+0.045, +0.055]$ & $1.1\!\cdot\!10^{-116}$ & $<2\!\cdot\!10^{-5}$ \\
        Qwen3-Emb-8B         & 4{,}999 & $+0.023$ & $[+0.018, +0.027]$ & $1.8\!\cdot\!10^{-24}$ & $<2\!\cdot\!10^{-5}$ \\
        Llama-Nemotron-8B    & 4{,}999 & $\mathbf{+0.051}$ & $[+0.046, +0.056]$ & $2.6\!\cdot\!10^{-192}$ & $<2\!\cdot\!10^{-5}$ \\
        Octen-Emb-8B         & 4{,}999 & $+0.027$ & $[+0.022, +0.031]$ & $1.3\!\cdot\!10^{-24}$ & $<2\!\cdot\!10^{-5}$ \\
        \teLargeShort{}      & 4{,}999 & $+0.040$ & $[+0.035, +0.044]$ & $1.1\!\cdot\!10^{-96}$ & $<2\!\cdot\!10^{-5}$ \\
        BM25 (Okapi)         & 4{,}999 & $+0.024$ & $[+0.020, +0.028]$ & $8.3\!\cdot\!10^{-50}$ & $<2\!\cdot\!10^{-5}$ \\
        \bottomrule
    \end{tabular}
    \caption{\hcmsynth{} paired dialect gap at $k{=}10$ (the contrast
    behind \cref{fig:aalwme-combined}a): mean per-sample $\drank =
    \mathrm{RR}_{\text{WME}} - \mathrm{RR}_{\text{AAL}}$, averaging the
    three synthetic AAL rewrites per WME source. Both paired tests
    (Wilcoxon, sign-flip permutation; \cref{app:sigtests}) reject on
    every retriever. Parallel to the \allsidesq{}
    \cref{tab:political-goldtest}.}
    \label{tab:hcmagic-real}
    \end{table*}

    \begin{table*}[t]\centering\small
    \setlength{\tabcolsep}{10pt}
    \begin{tabular}{lrrrr}
        \toprule
        Retriever & $n$ (WME/AAL) & MRR gap (WME$-$AAL) & 95\% CI & MWU $p$ \\
        \midrule
        BGE-large            & 659/647 & $+0.034$ & $[-0.010, +0.077]$ & 0.10 \\
        Qwen3-Emb-8B         & 659/647 & $\mathbf{+0.077}$ & $[+0.028, +0.125]$ & $2.3\!\cdot\!10^{-3}$ \\
        Llama-Nemotron-8B    & 659/647 & $\mathbf{+0.091}$ & $[+0.056, +0.125]$ & $8.7\!\cdot\!10^{-8}$ \\
        Octen-Emb-8B         & 659/647 & $+0.062$ & $[+0.015, +0.111]$ & 0.01 \\
        \teLargeShort{}      & 659/647 & $+0.064$ & $[+0.017, +0.111]$ & 0.01 \\
        BM25 (Okapi)         & 659/647 & $+0.040$ & $[+0.015, +0.065]$ & 0.03 \\
        \bottomrule
    \end{tabular}
    \caption{\hcmnat{} unpaired dialect gap at $k{=}10$ (the contrast
    behind \cref{fig:aalwme-combined}b): difference in mean MRR between
    the naturally-occurring WME and AAL query pools. Parallel to the
    \redditpol{} \cref{tab:reddit-goldtest}; bold marks the two largest
    gaps. BGE's CI includes $0$, consistent with its $p{=}0.10$.}
    \label{tab:hcmagic-nat-gap}
    \end{table*}

\section{Dialect k-sweep companions}
\label{app:aalwme-ksweeps}

    Companion k-sweep figures to
    \cref{fig:aalwme-combined}. The WME$>$AAL gap is stable
    across $k\in\{1, 5, 10, 20, 50, 100\}$ in both the paired \hcmsynth{}
    and the unpaired \hcmnat{} settings.

    \begin{figure}[h]
        \centering
        \includegraphics[width=0.95\linewidth]{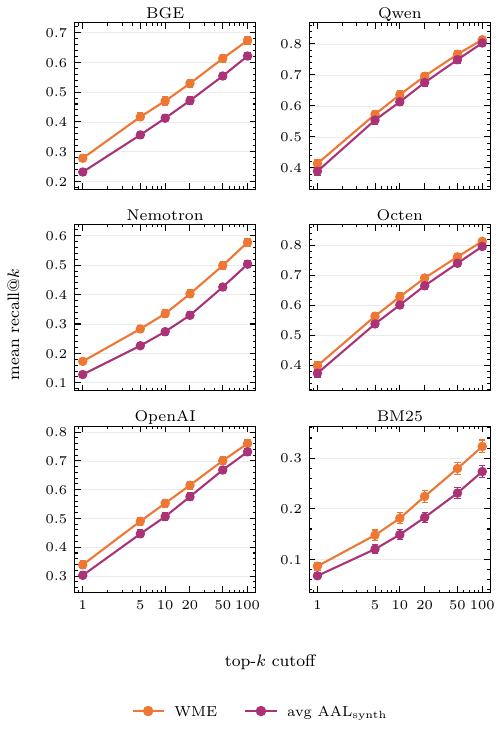}
        \caption{\hcmsynth{} mean Recall@$k$ across
        $k\in\{1, 5, 10, 20, 50, 100\}$, per retriever (per-panel
        y-zoom to make CIs visible at this $n$). Whiskers: 95\%
        bootstrap CIs.}
        \label{fig:aalwme-synth-ksweep}
    \end{figure}

    \begin{figure}[h]
        \centering
        \includegraphics[width=0.95\linewidth]{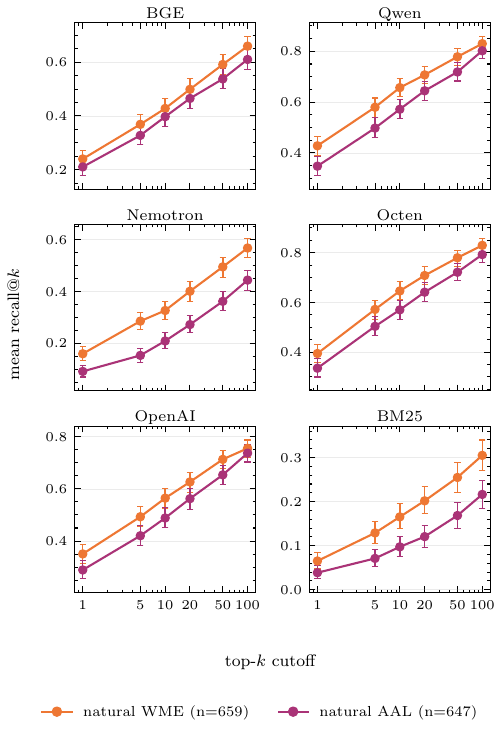}
        \caption{\hcmnat{} unpaired mean Recall@$k$ across
        $k\in\{1, 5, 10, 20, 50, 100\}$. Whiskers: 95\%
        bootstrap CIs.}
        \label{fig:aalwme-real-ksweep}
    \end{figure}

\begin{figure}[h]
        \centering
        \includegraphics[width=0.95\linewidth]{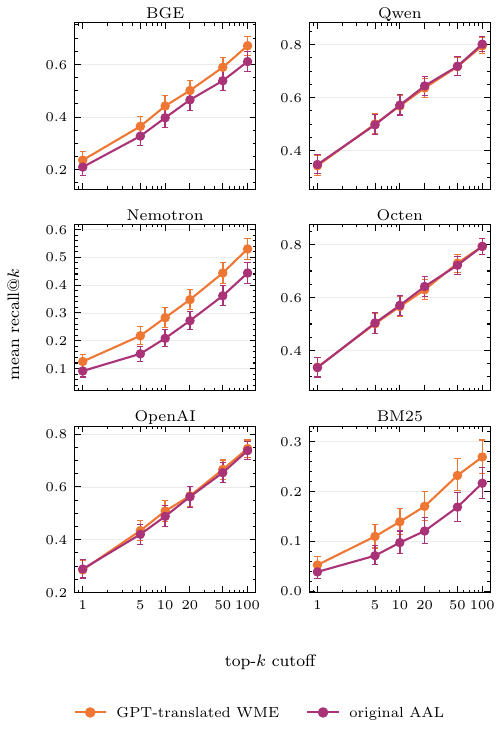}
        \caption{Translation-mitigation Recall@$k$ across
        $k\in\{1,5,10,20,50,100\}$ per retriever on the $n{=}647$
        paired \aalq{} queries. Whiskers: 95\% bootstrap CIs.
        Per-panel y-zoom to make CIs visible.}
        \label{fig:aalwme-translation-ksweep}
    \end{figure}

    \paragraph{Translation k-sweep.}
    Companion to \cref{fig:aalwme-translation-forest}: paired
    Recall@$k$ across $k\in\{1,5,10,20,50,100\}$ for GPT-translated
    \wmeq{} vs the original \aalq{} query on the same $n{=}647$
    \hcmnat{} \aalq{} queries (\cref{fig:aalwme-translation-ksweep}).
    The two-camp split visible in the main forest (Qwen / Octen /
    \teLargeShort{} overlap; BGE / Nemotron / BM25 separate) persists
    at every swept $k$.

\section{\hcmsynth{} augmentation quality}
\label{app:paraphrase-qc}

    The \hcmsynth{} synthetic AAL paraphrases are generated by the
    \citet{ziems-value} / \citet{deas-phonate} pipelines applied to
    HealthCareMagic patient queries. Both pipelines were validated
    with AAL-speaker reviewers on their own source corpora (TwitterAAL
    and LiveQA respectively) but were not independently re-validated
    on HealthCareMagic outputs. We report a quantitative + qualitative
    spot-check below.

    \paragraph{Quantitative QC (all $14{,}997$ paraphrases).}
    Median length ratio AAL/WME $= 1.06$; median character-level edit
    ratio (vs source WME) $= 0.58$ (5th percentile $= 0.20$, so even
    the least-shifted paraphrase contains substantial surface change);
    inter-paraphrase diversity (avg pairwise edit ratio across the
    three paraphrases per sample, $n{=}500$-sample subset) median $= 0.58$.
    Near-WME-duplicates (edit ratio $< 0.05$) total $19 / 14{,}997$
    ($0.13\%$); identical-to-WME $2 / 14{,}997$.

    \paragraph{Qualitative observations.}
    Random samples contain recognisable AAL features
    (phonological respellings, habitual aspect ``I been had'',
    ``y'all'', ``might can'' double modal) but also systematic
    generator artifacts: over-frequent cleft structures
    (``It is X that Y''), substitution of ``for me'' for ``my'',
    mid-sentence pronoun drift (``she/he'' replacing first-person
    ``I''), and occasional sub-word garbling (``Shems'' for ``I'm'',
    ``mat'' for ``might'').

    \paragraph{Example.}
    Source WME: \textit{``Hi I have a problem with my gum. Have a
    rednes on the left two front toths for about a year. have been at
    the dentist have my gums clean have an x ray atc. no one can help
    me. I also have venires done on my top. Can you please let me know
    what can cose infection.''}
    AAL paraphrase 1: \textit{``Hi {\color{blue}\textbf{She have}}
    a problem with {\color{blue}\textbf{for me}} gum. Do Have a rednes
    on the left two front toth for abat a year. have be at that
    dentist have my gum clean have an x ra atc. no one
    {\color{blue}\textbf{mat}} can help me. {\color{blue}\textbf{It is
    venires done on my top that i also has.}} Can
    {\color{blue}\textbf{yaw}} please let
    {\color{blue}\textbf{she}} know can what cos infection.''}
    (Bolded blue: real AAL feature \texttt{yaw} for \emph{y'all};
    generator artifacts \texttt{She have} pronoun drift,
    \texttt{for me} for \emph{my}, \texttt{mat can} as attempted
    double-modal \emph{might can}, and cleft restructure.) These
    artifacts likely inflate the \hcmsynth{} $\drank$ above what
    natural-AAL features alone would induce, particularly for
    out-of-distribution-sensitive strong encoders---the
    translation contrast (\cref{app:translation}) bounds the
    natural-AAL dialect penalty for the strong encoders at
    $\pm 0.015$ MRR.

\section{Translation-mitigation experiment}
\label{app:translation}

    A natural deployment fix for the AAL gap is to normalize the query into
    WME---the variety these retrievers handle best---before retrieving. We
    test this with \texttt{gpt-5-mini}: each of the $n{=}647$ \hcmnat{} AAL
    queries is back-translated to WME at $T{=}0$ and retrieved with the WME
    version. Because both forms share the same gold document, the contrast
    is paired (per-query $\drank = \mathrm{RR}_{\text{trans-WME}} -
    \mathrm{RR}_{\text{AAL}}$; paired Wilcoxon and sign-flip permutation).
    LLM translation has no content-fidelity guarantee and may add or drop
    medical content, so we report it as a practical intervention, not a
    clean dialect contrast; prior work also finds that LLMs struggle to
    rewrite and interpret AAL \citep{deas-aal, mckeown-aal}, making this a
    stringent test of the fix.

    \begin{figure}[t]
        \centering
        \includegraphics[width=0.95\linewidth]{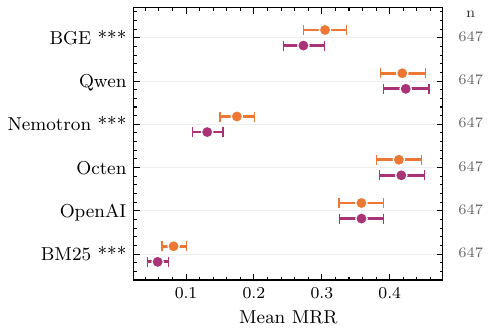}
        \caption{Translation-mitigation paired MRR on the $n{=}647$
        \hcmnat{} \aalq{} queries: original \aalq{} vs.\
        \texttt{gpt-5-mini} back-translation to \wmeq{}. Same content
        (modulo translation noise), tested with paired Wilcoxon.
        Translation $k$-sweep in \cref{app:aalwme-ksweeps}.}
        \label{fig:aalwme-translation-forest}
    \end{figure}

    Back-translation splits the retrievers cleanly
    (\cref{fig:aalwme-translation-forest},
    \cref{tab:hcmagic-trans-drank}): BGE-large, Llama-Nemotron-8B, and
    BM25 improve measurably ($p\!\approx\!0$), while Qwen3-Emb-8B,
    Octen-Emb-8B, and \teLargeShort{} do not change
    ($|\overline{\drank}|\leq 0.005$, all $p\!\geq\!0.22$). The retrievers
    that already handle AAL best benefit least, and---apart from BGE, which
    had no significant gap to begin with---no retriever closes the gap. A
    lexical-surface rewrite thus helps only the most dialect-sensitive
    retrievers and does not remove the disparity.

    \begin{table*}[t]\centering\small
    \setlength{\tabcolsep}{10pt}
    \resizebox{\linewidth}{!}{
        \begin{tabular}{lrrrrr}
            \toprule
            Retriever & $n$ pairs & $\overline{\drank}$ (trans-WME$-$AAL) & 95\% CI & Wilcoxon $p$ & Perm $p$ \\
            \midrule
            BGE-large            & 647 & $\mathbf{+0.032}$ & $[+0.015, +0.048]$ & $1.1\!\cdot\!10^{-7}$ & $2.0\!\cdot\!10^{-4}$ \\
            Qwen3-Emb-8B         & 647 & $-0.005$ & $[-0.020, +0.010]$ & 0.22 & 0.50 \\
            Llama-Nemotron-8B    & 647 & $\mathbf{+0.044}$ & $[+0.027, +0.061]$ & $8.5\!\cdot\!10^{-28}$ & $<2\!\cdot\!10^{-5}$ \\
            Octen-Emb-8B         & 647 & $-0.004$ & $[-0.019, +0.012]$ & 0.43 & 0.64 \\
            \teLargeShort{}      & 647 & $-0.000$ & $[-0.014, +0.014]$ & 0.89 & 0.99 \\
            BM25 (Okapi)         & 647 & $+0.023$ & $[+0.010, +0.037]$ & $1.2\!\cdot\!10^{-9}$ & $3.8\!\cdot\!10^{-4}$ \\
            \bottomrule
        \end{tabular}
    }
    \caption{Translation-mitigation paired gap at $k{=}10$: each \hcmnat{}
    AAL query back-translated to WME via \texttt{gpt-5-mini}, then $\drank =
    \mathrm{RR}_{\text{trans-WME}} - \mathrm{RR}_{\text{AAL}}$. Translation
    helps the dialect-sensitive retrievers (BGE, Llama-Nemotron, BM25:
    positive and significant) but not the rest (Qwen, Octen,
    \teLargeShort{}: CIs span $0$), and closes no gap fully.}
    \label{tab:hcmagic-trans-drank}
    \end{table*}

\section{Bias-gap method details}
\label{app:bias-gap-details}

    \paragraph{Tokenization for \lex{}.}
    Tokenization is a lowercase alphanumeric regex split with
    apostrophe-suffixes kept attached (so ``don't'' and ``she's'' are
    single tokens, punctuation dropped); we deliberately use a non-model
    tokenizer so the \lex{} channel is defined on the raw query text
    rather than on any retriever's BPE/wordpiece.

    \paragraph{Robust standard errors.}
    We use robust standard errors---specifically HC1
    heteroskedasticity-consistent, with cluster-robust adjustment within
    source article on \allsidesq{} and within source sample on \hcmsynth{};
    \redditpol{} and \hcmnat{} are unpaired so there is no within-cluster
    structure. The qualitative pattern in \cref{sec:bias-gap-results}
    holds under HC0--HC3.

\section{Lexical-residual regressions}
\label{app:lexreg}
\label{app:political-lexreg}
\label{app:dialect-lexreg}

    Per-retriever coefficients for the lexical-residual decomposition of
    \cref{sec:bias-gap} on all four datasets: the raw gap, the residual
    after partialling out the lexical channel ($\hat\alpha$ for the paired
    datasets, $\hat\gamma$ for the unpaired), and the lex slope
    $\hat\beta$.

    \paragraph{\allsidesq{} (paired).}
    Per-encoder OLS
    $\dlean = \alpha + \beta\,(\mathrm{lex}_R - \mathrm{lex}_L)_c$
    on each retriever's gold-filtered subset (cluster-robust SE on
    \texttt{article\_id}), where $(\mathrm{lex}_R - \mathrm{lex}_L)_c$
    is the Monroe--Colaresi--Quinn log-odds asymmetry between the right
    and left query of pair $c$ (mean $\approx -28$, SD $26$).
    \teLargeShort{} is the only retriever with a measurable
    slope; BM25 and the four open-weights have $\beta$
    indistinguishable from zero
    (\cref{tab:political-lexreg-gold}). After partialling out the lex
    channel, \teLargeShort{}'s residual $\alpha$ sits inside the
    open-weights' residual band, while BM25's residual has larger
    magnitude than every open-weights residual.

    \begin{table*}[h]\centering\small
    \setlength{\tabcolsep}{10pt}
    \begin{tabular}{lrrrr}
        \toprule
        Retriever & $n$ & $\overline{\dlean}$ & residual $\alpha$ & slope $\beta$ ($p$) \\
        \midrule
        BM25 (Okapi)         & 1{,}798 & $+0.043$ & $+0.038$ & $\approx 0\ (p{=}0.56)$ \\
        BGE-large            & 2{,}422 & $+0.027$ & $+0.027$ & $\approx 0\ (p{=}0.98)$ \\
        Qwen3-Emb-8B         & 2{,}508 & $+0.023$ & $+0.031$ & $+2.9\!\cdot\!10^{-4}\ (p{=}0.11)$ \\
        Llama-Nemotron-8B    & 1{,}416 & $+0.025$ & $+0.025$ & $\approx 0\ (p{=}0.94)$ \\
        Octen-Emb-8B         & 2{,}605 & $+0.023$ & $+0.018$ & $\approx 0\ (p{=}0.36)$ \\
        \teLargeShort{}      & 2{,}663 & $+0.045$ & $+0.029$ & $\mathbf{-5.7\!\cdot\!10^{-4}\ (p{=}0.002)}$ \\
        \bottomrule
    \end{tabular}
    \caption{Gold-filtered lex-residual fit. Each retriever's regression
    runs on its own gold-filtered subset (so $n$ varies; cross-retriever
    $\alpha$-comparison partly confounds with subset selection).
    \teLargeShort{} is the only retriever with a non-zero $\beta$; the
    lex channel accounts for $\overline{\dlean} - \hat\alpha = +0.016$ of
    its gap at the mean.}
    \label{tab:political-lexreg-gold}
    \end{table*}

    \paragraph{\redditpol{} (unpaired).}
    Per-encoder unpaired between-group regression at $k=20$ on
    \redditpol{} ($n_{L} = 1{,}327$ liberal-asker $+$ $n_{R} =
    1{,}225$ conservative-asker queries; centrist dropped):
    $\text{lean}_q = \alpha + \beta\,\text{lex}_q + \gamma\,\mathbb{1}[q \in \text{right}] + \varepsilon$
    with HC1 robust SE.
    The fit is over the full unfiltered top-$k$, so the raw gap
    here differs from the gold-filtered $k{=}10$ gaps of
    \cref{fig:political-combined}b.
    $\hat\gamma$ is the right$-$left lean gap after
    partialling out per-query MCQ asymmetry between left and right
    post-text vocabularies; $\hat\beta$ is the within-group lex slope.
    All five dense retrievers have positive significant $\hat\beta$
    (organic Reddit vocabulary is far more lexically partisan than
    \allsidesq{} query pairs); BGE's residual gap collapses to zero,
    Llama / Octen / \teLargeShort{} retain a significant non-lexical
    component, Qwen3 marginal.

    \begin{table*}[h]\centering\small
    \setlength{\tabcolsep}{8pt}
    \begin{tabular}{lrrrr}
        \toprule
        Retriever & raw R$-$L gap & $\hat\gamma$ ($p$) & $\hat\gamma / \text{raw}$ & $\hat\beta$ ($p$) \\
        \midrule
        BGE-large            & $+0.024$ & $+0.007\ (p{=}0.42)$           & $0.31$ & positive sig. \\
        Qwen3-Emb-8B         & $+0.028$ & $+0.015\ (p{=}0.08)$           & $0.55$ & positive sig. \\
        Llama-Nemotron-8B    & $+0.039$ & $\mathbf{+0.022\ (p{=}0.028)}$ & $0.55$ & positive sig. \\
        Octen-Emb-8B         & $+0.048$ & $\mathbf{+0.033\ (p<10^{-3})}$ & $0.70$ & positive sig. \\
        \teLargeShort{}      & $+0.062$ & $\mathbf{+0.036\ (p<10^{-3})}$ & $0.58$ & positive sig. \\
        BM25 (Okapi)         & $+0.001$ & $+0.002$ (NS)                  & --- & NS            \\
        \bottomrule
    \end{tabular}
    \caption{\redditpol{} unpaired lex-residual at $k=20$. Bolded
    $\hat\gamma$ are detectable encoder-level residual gaps after
    partialling out per-query MCQ asymmetry; BGE's collapse is
    consistent with topic-asymmetry exposure. $\hat\beta$ is positive
    and significant ($p<0.05$) for all five dense retrievers,
    contrasting with \allsidesq{} where only \teLargeShort{} had a
    detectable slope.}
    \label{tab:political-lexreg-realarm}
    \end{table*}

    \paragraph{\redditpol{}, gold-filtered outcome (robustness).}
    The fit above uses the unfiltered top-$k$; as a robustness check we
    repeat it with the outcome of \cref{fig:political-combined}b: per-query
    lean over the judged-relevant articles in the top-10 (both judges
    grade relevant, \cref{app:reddit-judge}), keeping queries with at
    least one relevant article. This decomposes the same quantity the
    main test reports, at the cost of smaller per-retriever $n$ (gold
    retention), a coarser per-query outcome (few relevant articles per
    query), and conditioning on judged relevance. The raw gaps reproduce
    the gold-filtered gaps of \cref{app:political-goldtest}, and the
    conclusions of the unfiltered fit carry over
    (\cref{tab:political-lexreg-realarm-gold}): BGE and Qwen3 show no
    detectable residual, while Llama-Nemotron and \teLargeShort{} retain
    large significant residuals ($\hat\gamma/\text{raw} \approx 0.8$);
    Octen is positive but not significant. BM25, which has no unfiltered
    gap, here carries the largest raw gap, and its residual is large but
    marginal ($p{=}0.07$). $\hat\beta$ is significant for BGE
    ($p{<}10^{-4}$), Qwen3 ($p{=}0.008$), and Octen ($p{=}0.03$), and not
    for the others.

    \begin{table*}[h]\centering\small
    \setlength{\tabcolsep}{8pt}
    \begin{tabular}{lrrrr}
        \toprule
        Retriever & $n$ (lib/con) & raw R$-$L gap & $\hat\gamma$ ($p$) & $\hat\gamma / \text{raw}$ \\
        \midrule
        BGE-large            & $616/474$ & $+0.109$ & $+0.020\ (p{=}0.67)$           & $0.18$ \\
        Qwen3-Emb-8B         & $698/539$ & $+0.069$ & $+0.017\ (p{=}0.69)$           & $0.24$ \\
        Llama-Nemotron-8B    & $432/317$ & $+0.186$ & $\mathbf{+0.148\ (p{=}0.017)}$ & $0.79$ \\
        Octen-Emb-8B         & $729/553$ & $+0.103$ & $+0.061\ (p{=}0.13)$           & $0.59$ \\
        \teLargeShort{}      & $747/570$ & $+0.136$ & $\mathbf{+0.106\ (p{=}0.009)}$ & $0.78$ \\
        BM25 (Okapi)         & $284/228$ & $+0.195$ & $+0.138\ (p{=}0.07)$           & $0.71$ \\
        \bottomrule
    \end{tabular}
    \caption{\redditpol{} lex-residual with the gold-filtered
    relevant-subset lean@10 of \cref{fig:political-combined}b as the
    outcome, on queries with $\geq 1$ both-judge-relevant article.
    Bolded $\hat\gamma$ are detectable residual gaps after partialling
    out per-query MCQ asymmetry.}
    \label{tab:political-lexreg-realarm-gold}
    \end{table*}

    \paragraph{\hcmsynth{} (paired).}
    Per-encoder OLS
    $\drank = \alpha + \beta\,\text{asym}_r + \varepsilon_r$
    on all $4{,}999$ paired samples with HC1 robust SE, where
    $\text{asym}_r = \mathrm{lex}_{\text{WME}_r} -
    \overline{\mathrm{lex}}_{\text{AAL}_{\text{synth}_r}}$ is the
    per-sample Monroe--Colaresi--Quinn log-odds asymmetry summed over
    tokens (mean $\overline{\text{asym}} = 574$, SD $370$). $\zeta$
    scores are computed against the union vocabulary of all $4{,}999$
    WME inputs ($\sim\!425$k tokens) vs.\ all $14{,}997$ AAL paraphrases
    ($\sim\!1.34$M tokens) with the same informative-Dirichlet prior
    ($\alpha_0 = 0.01$) as the political fit
    (\cref{tab:political-lexreg-gold}).

    \begin{table*}[h]\centering\small
    \setlength{\tabcolsep}{8pt}
    \begin{tabular}{lrrrrr}
        \toprule
        Retriever & $n$ & $\overline{\drank}$ & residual $\alpha$ & slope $\beta$ ($p$) & $\beta\!\cdot\!\overline{\text{asym}}$ \\
        \midrule
        Llama-Nemotron-8B    & 4{,}999 & $+0.051$ & $+0.046$ & $+8.4\!\cdot\!10^{-6}\ (p{=}0.21)$ & $+0.005$ \\
        BGE-large            & 4{,}999 & $+0.050$ & $+0.047$ & $+5.8\!\cdot\!10^{-6}\ (p{=}0.43)$ & $+0.003$ \\
        \teLargeShort{}      & 4{,}999 & $+0.040$ & $+0.039$ & $+5.9\!\cdot\!10^{-7}\ (p{=}0.93)$ & $+0.000$ \\
        Octen-Emb-8B         & 4{,}999 & $+0.027$ & $+0.024$ & $+4.8\!\cdot\!10^{-6}\ (p{=}0.48)$ & $+0.003$ \\
        BM25 (Okapi)         & 4{,}999 & $+0.024$ & $+0.024$ & $+4.2\!\cdot\!10^{-7}\ (p{=}0.93)$ & $+0.000$ \\
        Qwen3-Emb-8B         & 4{,}999 & $+0.023$ & $+0.021$ & $+2.7\!\cdot\!10^{-6}\ (p{=}0.68)$ & $+0.002$ \\
        \bottomrule
    \end{tabular}
    \caption{Per-sample lex-residual fit on \hcmsynth{}
    (sorted by raw $\overline{\drank}$). All $\hat\beta$ point estimates
    are positive (consistent direction) but none are statistically
    distinguishable from zero; the implied per-sample lex contribution
    $\beta\!\cdot\!\overline{\text{asym}}$ is at most $0.005$ MRR
    (under $10\%$ of the raw paired gap for the open-weights, $\approx 0$
    for \teLargeShort{} and BM25). Residual $\alpha$ ranking matches
    the raw $\overline{\drank}$ ranking up to two micro-swaps within
    noise (BGE/Llama by $10^{-3}$; BM25/Octen by $2\!\cdot\!10^{-6}$).}
    \label{tab:dialect-lexreg}
    \end{table*}

    \paragraph{\hcmnat{} (unpaired).}
    Per-encoder unpaired between-group regression on \hcmnat{}
    ($647$ natural AAL $+$ $659$ natural WME queries):
    $\text{MRR}_q = \alpha + \beta\,\text{lex}_q + \gamma\,\mathbb{1}[q \in \text{WME}] + \varepsilon$
    with HC1 robust SE. $\hat\gamma$ is the WME$-$AAL gap after
    partialling out per-query MCQ asymmetry between AAL and WME
    post-text vocabularies. No retriever has a detectable
    within-group lex slope (smallest $\hat\beta$ $p = 0.15$);
    $\hat\gamma$ is detectable on five of six retrievers (BGE
    marginal at $p = 0.21$); the slight excess of $\hat\gamma$ over the
    raw gap on five of six arises because $\hat\beta$ is mildly negative
    while the WME pool has higher mean lex. Note the caveat in
    \cref{sec:aalwme-lexreg-main}: AAL-distinctive tokens include
    content words (\textit{she}, \textit{her}, \textit{periods},
    \textit{baby}), so lex-residual partials out token-frequency
    asymmetry but not content-difficulty.

    \begin{table*}[h]\centering\small
    \setlength{\tabcolsep}{8pt}
    \begin{tabular}{lrrrr}
        \toprule
        Retriever & raw WME$-$AAL gap & $\hat\gamma$ ($p$) & $\hat\gamma / \text{raw}$ & $\hat\beta$ ($p$) \\
        \midrule
        Llama-Nemotron-8B    & $+0.091$ & $\mathbf{+0.085\ (p{=}0.0018)}$ & $0.93$ & NS \\
        Qwen3-Emb-8B         & $+0.077$ & $\mathbf{+0.110\ (p{=}0.0018)}$ & $1.42$ & NS \\
        \teLargeShort{}      & $+0.064$ & $\mathbf{+0.081\ (p{=}0.023)}$  & $1.26$ & NS \\
        Octen-Emb-8B         & $+0.062$ & $\mathbf{+0.098\ (p{=}0.0049)}$ & $1.59$ & NS \\
        BM25 (Okapi)         & $+0.040$ & $\mathbf{+0.041\ (p{=}0.043)}$  & $1.01$ & NS \\
        BGE-large            & $+0.034$ & $+0.042\ (p{=}0.21)$            & $1.21$ & NS \\
        \bottomrule
    \end{tabular}
    \caption{\hcmnat{} unpaired lex-residual. Bolded $\hat\gamma$ are
    significant at $p < 0.05$. The residual gap survives partialling
    on every retriever with $\hat\gamma / (\text{raw gap})
    \in [0.93,\, 1.59]$; no retriever has a detectable within-group
    lex slope. Conclusion qualitatively matches the \hcmsynth{} result
    in \cref{tab:dialect-lexreg}.}
    \label{tab:dialect-lexreg-realarm}
    \end{table*}

\section{Probing Details}
\label{app:probing-dets}

    \paragraph{Extracting Hidden Representations. } For encoder-only models, we use mean pooling on the outputs of each layer to represent hidden representations. We additionally apply standard scaling.
    \teLargeShort{} is excluded from probing, as the API exposes no intermediate layers.

    \paragraph{Probe Training. } We train linear probes using \texttt{LogisticRegression} models from \texttt{sklearn} \cite{scikit-learn}. We use the default hyperparameters and set the maximum iterations to 5000 to allow for convergence.

\end{document}